\documentclass[]{achemso}
\setkeys{acs}{articletitle=true}
\setkeys{acs}{etalmode=truncate,maxauthors=10}
\usepackage[version=3]{mhchem} %
\usepackage{color}
\usepackage{MnSymbol}
\usepackage{booktabs}
\usepackage{multirow}
\usepackage{mathdots}
\usepackage{xcolor}
\usepackage{comment}
\usepackage{tabularx}
\usepackage{bm}
\usepackage{xr}
\usepackage{todonotes}
\usepackage[symbol]{footmisc}
\mciteErrorOnUnknownfalse
\usepackage{xr}
\makeatletter

\SectionNumbersOn



\author{Lauri Seppäläinen}
\affiliation[University of Helsinki]
{Department of Computer Science, Pietari Kalmin katu 5, University of Helsinki, Helsinki, FI 00560}
\email{lauri.seppalainen@helsinki.fi}
\author{Jakub Kube{\v c}ka}
\affiliation[Aarhus University]
{Department of Chemistry, Aarhus University, Langelandsgade 140, Aarhus C, DK 8000}
\author{Jonas Elm}
\affiliation[Aarhus University]
{Department of Chemistry, Aarhus University, Langelandsgade 140, Aarhus C, DK 8000}
\author{Kai R. Puolamäki}
\affiliation[University of Helsinki]
{Department of Computer Science, Pietari Kalmin katu 5, University of Helsinki, Helsinki, FI 00560}

\title{Fast and Interpretable Machine Learning Modelling of Atmospheric Molecular Clusters}

\begin{document}
\newpage
\begin{abstract}
Understanding how atmospheric molecular clusters form and grow is key to resolving one of the biggest uncertainties in climate modelling: the formation of new aerosol particles. 
While quantum chemistry offers accurate insights into these early-stage clusters, its steep computational costs limit large-scale exploration. 
In this work, we present a fast, interpretable, and surprisingly powerful alternative: \textit{k}-nearest neighbour (\textit{k}-NN) regression model. 
By leveraging chemically informed distance metrics---including a kernel-induced metric and one learned via metric learning for kernel regression (MLKR)---we show that simple \textit{k}-NN models can rival more complex kernel ridge regression (KRR) models in accuracy, while reducing computational time by orders of magnitude. 
We perform this comparison with the well-established Faber--Christensen--Huang--Lilienfeld (FCHL19) molecular descriptor, but other descriptors (e.g., FCHL18, MBDF, and CM) can be shown to have similar performance.
Applied to both simple organic molecules in the QM9 benchmark set and large datasets of atmospheric molecular clusters (sulphuric acid--water and sulphuric--multibase-base systems), our \textit{k}-NN models achieve near-chemical accuracy, scale seamlessly to datasets with over 250,000 entries, and even appears to extrapolate to larger unseen clusters with minimal error (often nearing 1 kcal/mol). 
With built-in interpretability and straightforward uncertainty estimation, this work positions \textit{k}-NN as a potent tool for accelerating discovery in atmospheric chemistry and beyond.
\end{abstract}

\newpage
\section{Introduction}\label{sec:intro}

New particle formation (NPF) is the dominant source of aerosol particles in the atmosphere.\cite{kulmala25,zhao2024global,cai24}
NPF contributes substantially to the global cloud condensation nuclei (CCN) budget, with 10--80\% of the number concentration, depending on region.\cite{CCN_budget,CCN,Trostl_nature_2016}
As specified by the most recent IPCC report\cite{IPCC2021new}, the largest source of uncertainty in estimating Earth's current and future radiative balance originates from our lack of understanding of how aerosol particles are formed and how many of them eventually reach CCN sizes of roughly 50--100~nm.

The NPF process is initiated by the formation of strongly bound atmospheric molecular clusters \cite{kulmala2013direct}.
Measuring the cluster formation process using current experimental instrumentation is challenging, as soft ionization mass spectrometer techniques are not able to identify all clusters simultaneously.\cite{lee19}
Accurate quantum chemical (QC) calculations can capture clustering of single molecules up to small 1--2~nm particle sizes.
However, the desired accuracy comes at a cost of steep computational scaling with the studied system size.
For instance, the density functional theory methods typically employed for obtaining cluster structures scale roughly as $\mathcal{O}(N^4)$, where $N$ represents the number of basis functions or orbitals, and highly accurate CCSD(T) methods scale roughly as $\mathcal{O}(N^7)$.
Localization approaches such as DLPNO-CCSD(T)\cite{riplinger13a,riplinger13b,riplinger16}, PNO-CCSD(T)\cite{tew19,schmitz18}, or LNO-CCSD(T)\cite{nagy2018,nagy19,kallay2020,mrcc} can bring the scaling down, but still remain expensive on large cluster systems.

A promising strategy to overcome the computational costs of quantum chemical methods is the application of machine learning (ML). 
ML methods have been widely applied in chemistry as powerful tools in a broad spectrum of tasks, including drug discovery \cite{smith2018transforming}, reaction pathway discovery \cite{baiardi2022expansive}, estimation of material properties \cite{verma2023application}, and the analysis of complex molecular datasets generated by experimental techniques.\cite{Bortolussi25,Baird23,Paritosh22}
Due to their relatively low marginal computational cost, machine learning methods can be trained with large datasets and can hence rival the performance of QC calculations.
Furthermore, the inference cost, i.e., applying machine learning methods to novel datapoints, is often significantly lower than the training cost.
Thus, machine learning methods are increasingly utilised for interpolating results from QC calculations, as full replacements for these calculations, as well as pre-screening tools to choose the most promising candidate structures for more accurate calculations.\cite{keith21,meuwly21,kuntz22,katritzky10}
However, the computational costs for ML methods cannot be entirely ignored either.
These costs can limit the applicability of ML, particularly with large datasets, alongside the lack of mature, user-friendly software \cite{anstine2023machine}.
Also, estimating prediction uncertainties and extrapolation to outside training data are non-trivial problems without off-the-shelf solutions, even with ML.

Due to the lack of appropriate databases, ML has only been scarcely applied to atmospheric chemistry.\cite{sandstrom25}
In recent years, however, a variety of ML techniques have been explored for predicting key atmospheric properties.
For instance, molecular saturation vapour pressures modelled via kernel ridge regression (KRR\cite{lumiaro21}), extreme minimal learning machine (EMLM\cite{hyttinen22}), and neural networks (NN\cite{kruger25}) have shown promise.
Neural networks, in particular, have become a popular choice across chemistry for their flexibility and ability to model highly non-linear relationships.
Recent work has demonstrated their ability to predict binding energies of atmospheric clusters with impressive accuracy.\cite{kubecka2024accurate,gupta24, jiang22, jiang23} 
However, the benefits of NNs often come at the cost of interpretability and computational complexity, especially during training.
In many practical scenarios, neural networks require significantly larger datasets to avoid overfitting and can be prone to unpredictable generalization behaviour outside the training domain.
Additionally, their ``black-box'' nature may limit chemical insight---an important factor when trying to understand the physical principles driving atmospheric cluster formation.

In contrast, kernel-based models such as KRR offer a compelling balance between accuracy and transparency.
Kube{\v c}ka et al. introduced KRR combined with $\Delta$-learning\cite{Ramakrishnan_jctc_2015} to predict cluster binding energies of acid--base clusters with sub-1~kcal/mol accuracy.\cite{kubecka2022Quantum}
In addition, Knattrup et al. demonstrated that KRR can even be used to extrapolate beyond the training database regarding system size.\cite{knattrup2023clusterome}
However, as the reference database increases in size beyond $10^4$ data points, both training and inference times become slow, making KRR less practical for large datasets.

\begin{figure}
    \centering
    \includegraphics[width=\linewidth]{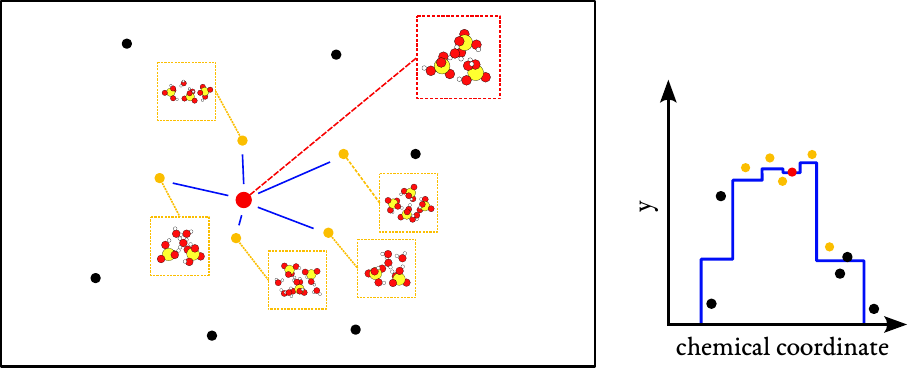}
    \caption{A simplified illustration of a $k$-NN model on chemical data. The prediction for the red structure is formed based on the $k=5$ nearest neighbours (yellow). The left panel shows a two-dimensional embedding of a chemical space with a test point in red, nearest neighbours in yellow and other training datapoints in black. The insets show example sulphuric acid--water clusters. The right panel shows how the $k$-NN regression prediction might behave along a one-dimensional line in the chemical space.}
    \label{fig:abstract}
\end{figure}

The success of KRR in modelling chemical systems implies that other instance- or similarity-based models may also perform well.
Perhaps the simplest instance-based model is the well-studied $k$-nearest neighbours ($k$-NN) regression model, where a prediction for a novel item is formed as a (weighted) average of the labels of the $k$ nearest datapoints in the training set \cite{hastie2009}, as demonstrated in Fig.~\ref{fig:abstract}.
Using tree-based data structures, the construction of the data structure (``training'' the $k$-NN model) will take $\mathcal{O}(pn(\log n)^2)$ time and the cost of inference (``prediction'') can be made to scale as $\mathcal{O}(k\log n)$, where $n$ is the number of training datapoints, $k$ is the number of neighbours, and $p$ is the dimensionality of the data vectors (e.g., cluster descriptors) \cite{omohundro1989Five}.
$k$-NN is therefore expected to incur much lower computational costs than KRR as the dataset size increases.
Besides increased speed, a $k$-NN model is also readily interpretable; the user can inspect the nearest neighbours (or their 3D structure) to understand how the model produces predictions.
The primary challenge in employing a $k$-NN model is determining an effective distance metric in high-dimensional space.
Without careful consideration for the choice of metric, the notion of ``nearest neighbours" can become meaningless with as few as 10--15 features. \cite{beyer1999meaningful}

In this paper, we address this challenge by employing metric learning or, as an alternative option, deriving a distance measure from a kernel function that has been shown to perform well in conjunction with a KRR model.
Our main contributions are to describe and evaluate several $k$-nearest neighbour modelling strategies for predicting the properties of chemical systems.
We demonstrate that our methods yield fast and interpretable predictions for the chemical properties of acid--base clusters, with a computational cost orders of magnitude smaller than conventional machine learning methods, at only a slight cost in prediction accuracy.
Furthermore, our proposed method is interpretable and readily allows for uncertainty estimation.

\section{Theory}\label{sec:theory}

In this section, we first define the machine learning problem we aim to solve in concrete terms.
We then describe KRR and $k$-NN models and the specific algorithms we use to mitigate the curse of dimensionality in $k$-NN.
Finally, we briefly introduce the FCHL descriptor we use to convert 3D structures of chemical systems to training data suitable for the machine learning models.

\subsection{Problem definition}

Let $\mathcal{D} = \{(\bm{x}_1, y_1), \ldots, (\bm{x}_n, y_n)\}$ be a dataset consisting of datapoints $\bm{x}_i$ and labels $y_i$, where $i\in \{1, ..., n\}$.
The datapoints $\bm{x}_i$ can be, e.g., molecular descriptors (see examples in Section~\ref{ssec:descriptors}).
The labels correspond to some chemical property of interest, such as the electronic binding energy.
In machine learning, we aim to find the function $f(\bm{x})$ which minimises the value of a chosen loss function $\mathcal{L}(y_i', f(\bm{x}_i'))$.
In this paper, we use the mean absolute error (MAE) as the loss function for evaluation:
\begin{equation}
    \mathcal{L}(y'_i, f(\bm{x}_i')) = \frac{1}{n} \sum\nolimits_{i=1}^{n} \lvert y'_i - f(\bm{x}_i') \rvert.
\end{equation}
Unless otherwise specified, we present the losses on a separate test (validation) set, which is not used in training.

\subsection{Instance-based models}\label{ssec:instance_based}
Machine learning methods can be divided into two categories: parametric and non-parametric models \cite{murphy2022probabilistic}.
In parametric models, the aim is to learn a set of parameters $\theta$ for a fixed-form function $f$ by minimising the loss on a dataset $\mathcal{D}$.
Examples include ordinary least squares linear regression and the more complex, and increasingly ubiquitous, neural networks.
Non-parametric models, on the other hand, assume no fixed form for $f$.
In this paper, we focus on instance-based non-parametric models, which learn by ``memorising'' the training data and produce predictions by leveraging some similarity metric between the training data and novel datapoints.
Such instance-based models can be expressed as
\begin{equation}
    f(\bm{x}') = \sum\nolimits_{j \in J} w(\bm{x}', \bm{x}_j) y_j,
    \label{eq:instance}
\end{equation}
where $w(\bm{x}', \bm{x}_j)$ is a weighting function between the novel item $\bm{x}'$ and datapoints $\bm{x}_j$ in a (sub)set $J \subseteq \mathcal{D}$ of training data.
Examples of the weighting function include similarity- or distance-based functions such as kernel functions.

In computational chemistry, one of the commonly used instance-based models is kernel ridge regression (KRR) \cite{janet2020Machinea}.
As the name suggests, KRR builds on ridge regression, a classic statistical model that incorporates a quadratic penalty into the linear regression parameters to mitigate overfitting.
KRR increases the model flexibility by first casting the input datapoints $\bm{x}$ to a feature space, denoted by $\phi(\bm{x})$. 
Usually, $\phi$ is a non-linear function and $\phi({\bm{x}})$ has higher dimensionality than the original input space.
Due to the ``kernel trick,'' the datapoints do not explicitly need to be projected into the feature space; merely defining the dot product between datapoints in this space suffices.
The dot product function $k(\bm{x}, \bm{x}')=\phi(\bm{x})\cdot\phi(\bm{x}')$ is referred to as the kernel, and it measures the similarity between datapoints in the feature space.
Now the prediction from a KRR model can be expressed as a weighted sum of the training labels:
\begin{equation}
    f_{\textrm{KRR}}(\bm{x}) = \sum\nolimits_{i=1}^n \alpha_i k(\bm{x}_i, \bm{x}) 
\end{equation}
from which the optimal weights can be solved from the following matrix equation:
\begin{equation}
    \bm{\alpha} = (\bm{K}+\lambda \bm{I}_n)^{-1}\bm{y}.
    \label{eq:krr_alpha}
\end{equation}
Here, the kernel matrix $\bm{K}$ consists of kernel evaluations between training datapoints: $\bm{K}_{i,j} = k(\bm{x}_i, \bm{x}_j)$, $\lambda$ is the ridge regularisation penalty and $\bm{I}_n$ is a $n \times n$ identity matrix.
In the terms of Eq.~\eqref{eq:instance}, the set $J$ is the full training dataset and the weighting function is $w(\bm{x}', \bm{x}_j) = (\bm{K} + \lambda \bm{I}_n)^{-1}k(\bm{x}', \bm{x}_j)$.
Computationally, finding $\bm{\alpha}$ by solving Eq.~\eqref{eq:krr_alpha} is expensive as it involves inverting the $n \times n$ kernel matrix $\bm{K} + \lambda\bm{I}_n$, which is an $\mathcal{O}(n^3)$ operation in practice \cite{golub2013matrix}, even though $\mathcal{O}(n^{2.376})$ matrix inversion algorithms exist \cite{coppersmith1990Matrix}.
Furthermore, while training the model incurs this cost only once, inference for $m$ samples is $\mathcal{O}(mn)$, which can become prohibitively expensive for large $n$, especially if evaluating the kernel function $k$ is non-trivial.

Properties of chemical systems can be divided into \emph{extensive} properties, which are size- or scale-dependent, and \emph{intensive} properties, which are size-independent.
In this paper, we focus on electronic binding energies, which are an extensive property; larger systems generally have lower (more negative) electronic binding energies.
One approach to model extensive properties with KRR is to assume that the extensive property can be decomposed into a sum of atomic contributions.
The formulation of KRR in this case stays the same, save for rewriting the kernel as a sum of pairwise atomic kernels \cite{langer2022Representations}.

Accuracy of ML models can be further enhanced by using $\Delta$-learning, a hybrid approach combining machine learning with fast quantum chemical calculations. \cite{ramakrishnan2015big}
In $\Delta$-learning the aim is not to directly predict a property, such as electronic binding energy, but instead a correction term between, for example, a related quantity (often more easily computed) and the target property (such as between energy and enthalpy), between different geometries (such as isomers of the same compound), or between two quantum chemical properties obtained at different levels of theory.
In a previous work, Kube{\v c}ka \& al. have shown how a $\Delta$-learning approach with KRR can achieve results within chemical accuracy when applied to atmospheric cluster data.\cite{kubecka2022Quantum}
$\Delta$-learning is not limited to KRR, but can be applied in a wide range of ML methods.
Additionally, the approach can also be used to optimise computational cost.
When the correction term ($\Delta$) represents the difference between two quantum chemical methods---or between a target property and a related, more easily computed quantity---$\Delta$-learning allows predictions from a simpler or less accurate model to be adjusted toward those of a more accurate, higher-level method. 
This can significantly improve accuracy at a marginal computational cost increase.

In contrast to a KRR model, in a $k$-NN model, each prediction for a novel item is simply the (weighted or unweighted) average of the labels of the $k$ nearest neighbours of that item in the training set:
\begin{equation}
    f_{\textrm{k-NN}}(\bm{x}) = \frac{1}{k} \sum\nolimits_{i \in \mathcal{N}_k(\bm{x}, d)} w_i({\bm{x}}) y_i,
\end{equation}
where $\mathcal{N}_k(\bm{x}, d)$ denotes the set of $k$ nearest neighbours for the item $\bm{x}$ as per the distance measure $d(\bm{x}, \bm{x}_i)$.
In the unweighted variant, the weight vector is a unit vector ($\bm{w} = \bm{1}$) and the prediction is the mean of the nearest labels.
However, if the labels can be assumed to vary smoothly with increasing distance, the accuracy of the predictions can be increased by weighing the labels with the reciprocal of the distance: $w_i = 1/d(\bm{x}, \bm{x}_i)$.
These two are the standard choices for weighting and are hence what we study in this paper.
The simplest $k$-NN training consists of simply memorizing the training samples and hence has an $\mathcal{O}(np)$ complexity, where $p$ is the dimensionality of the data vectors.
Inference for a novel item is $\mathcal{O}(knp)$ for a naïve implementation.
However, with a tree-based data structure, inference complexity can be pushed down to $\mathcal{O}(k\log n)$ with slightly higher training cost of $\mathcal{O}(pn(\log n)^2)$ needed for building the tree data structure used to find the nearest neighbours efficiently \cite{omohundro1989Five}.

\subsection{$k$-NN algorithms}

\begin{figure}[t]
    \centering
    \includegraphics[width=\linewidth]{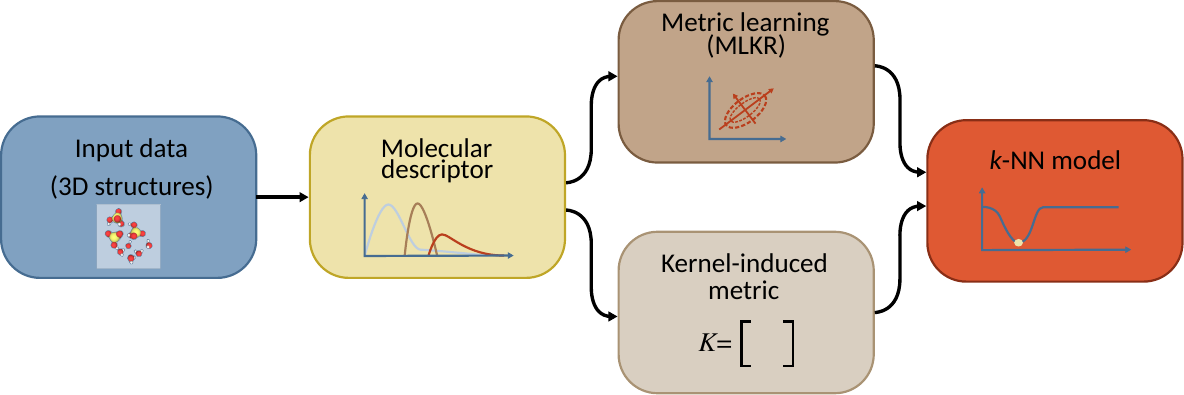}
    \caption{Schematic of the $k$-NN pipeline. The raw data consists of atomic positions in 3D space, from which we derive molecular descriptors (representations). Then, using one of two different distance measures, a learned one based on the MLKR algorithm and a metric derived from the KRR kernel, we construct the final $k$-NN model.}
    \label{fig:knn_pipeline}
\end{figure}

The main issue limiting the applicability of nearest neighbour models is the curse of dimensionality.
As the dimensionality of the input space increases, the distinction between the Euclidean distances of datapoints becomes less and less pronounced; the higher the dimensionality, the more equal the distances typically become.\cite{beyer1999meaningful}
If a large subset of datapoints is nearly the same distance away from the query item, it is difficult to meaningfully distinguish the nearest neighbours.
Hence, the key to applying $k$-NN to high-dimensional data, such as molecular descriptors, is a well-specified distance metric.
We employed two approaches to finding a suitable distance metric for comparison: kernel-induced distance and metric learning.
The two approaches are depicted in the $k$-NN modelling pipeline in Fig.~\ref{fig:knn_pipeline}.

KRR with the Faber--Christensen--Huang--Lilienfeld (FCHL) representation has shown promising results in the past \cite{kubecka2022Quantum}, and the key property for this performance is the use of a kernel function $k$ that can meaningfully measure the similarity between chemical systems.
If we have access to a good similarity metric, a natural assumption is that we can construct a distance metric from it.
This is indeed the case, provided certain assumptions are made about the properties of the kernel.
The \emph{kernel-induced metric} \cite{phillips2011gentle} is defined as:
\begin{equation}
    d(\bm{x}_i, \bm{x}_j) = \sqrt{k(\bm{x}_i,\bm{x}_i) + k(\bm{x}_j,\bm{x}_j) - 2 k(\bm{x}_i, \bm{x}_j)},
    \label{eq:kid}
\end{equation}
i.e., as the difference between self-similarities and cross-similarity of datapoints $i$ and $j$.
This definition is just a reformulation of the distance between the datapoints in the high-dimensional space:
\begin{equation}
||\phi_i - \phi_j||^2 = (\phi_i - \phi_j)\cdot(\phi_i - \phi_j)=\phi_i\cdot\phi_i+\phi_j\cdot\phi_j-2\,\phi_i\cdot\phi_j,
\end{equation}
written by using definition of the kernel as the dot product in the high-dimensional space $\phi({\bf x})$ of datapoint ${\bf x}$, i.e., $k(\bm{x}, \bm{x}')=\phi({\bf x})\cdot\phi({\bf x}')$.
The distance function $d$ is a valid pseudometric if $k$ is a positive definite kernel, which is usually assumed and is the case for the kernels used in this manuscript.
The FCHL kernel is a sum of pairwise atomic radial basis function (RBF) kernels \cite{faber2018alchemical}, which are positive definite, and it can be trivially shown that a sum of positive definite matrices is also a positive definite matrix.
Hence, when using the FCHL kernel, the induced kernel distance $d$ is a valid distance metric and can be used to create a $k$-NN model.
However, when using KRR to model an extensive property with a kernel composed of a sum of pairwise atomic kernels, the kernel is not normalised, meaning that the diagonal elements are not equal to 1.
As we will discuss in the Section \ref{sec:results}, this may lead to a decrease in performance compared to a corresponding KRR model.
If the self-similarity terms $k(\bm{x}_i, \bm{x}_i)$ and $k(\bm{x}_j, \bm{x}_j)$ in Eq.~\eqref{eq:kid} are high in the extensive kernel, the least dissimilar items based on the kernel-induced distance will be different than the most similar items based on the kernel function, causing the kernel-based $k$-NN and KRR models to produce different results.
This could, in principle, be mitigated via normalising the kernel such that the diagonal elements are always equal to 1, but this normalisation loses the extensiveness property of the kernel and can hence also decrease accuracy.

The other option we considered is metric learning.
As the name suggests, the goal of this approach is to learn a distance metric that can distinguish between similar and dissimilar datapoints in a meaningful way regarding a specific learning goal.
In regression, a natural choice is to find a distance metric such that datapoints with similar labels have short distances between each other and vice versa for datapoints with dissimilar labels.
In this paper, we employ the Metric Learning for Kernel Regression (MLKR) algorithm \cite{weinberger2007metric} as our metric learning algorithm of choice.
Originally developed for kernel regression, the goal of MLKR is to learn a Mahalanobis metric that minimizes the leave-one-out quadratic regression error.
The objective function to minimise is defined as
\begin{equation}
    \mathcal{L} = \sum\nolimits_{i=1}^n (y_i - f_{\textrm{LOO-KR}}(\bm{x}_i))^2,
\end{equation}
where $f_{\textrm{LOO-KR}}$ is the leave-one-out kernel regression (LOO-KR) model, defined as
\begin{equation}
    f_{\textrm{LOO-KR}}(\bm{x}_i) = \frac{\sum\nolimits_{j \neq i}y_j k(\bm{x}_i, \bm{x}_j)}{\sum\nolimits_{j \neq i}k(\bm{x}_i, \bm{x}_j)}
\end{equation}
and 
\begin{equation}
    k(\bm{x}_i, \bm{x}_j) = \frac{1}{\sigma \sqrt{2 \pi}} \exp \left(- \frac{d_{\bm{M}}(\bm{x}_i, \bm{x}_j)}{\sigma^2}\right)
\end{equation}
is the standard RBF kernel.
The learnable parameter in MLKR is the positive semidefinite matrix $\bm{M}$ that defines the Mahalanobis distance $d_{\bm{M}}$:
\begin{equation}
    d_{\bm{M}}(\bm{x}_i, \bm{x}_j) = (\bm{x}_i - \bm{x}_j)^T \bm{M} (\bm{x}_i - \bm{x}_j).
\end{equation}
The Mahalanobis distance is a generalization of the Euclidean distance metric, with an added linear transformation of the input space.
For efficiency, $\bm{M}$ can be decomposed into a matrix product
($\bm{M} = \bm{A}^\intercal\bm{A}$)
to ensure positive semidefiniteness without expensive checks after each optimisation step.
Then the distance measure can be trained using standard gradient-based optimisation methods with the following explicit gradient:
\begin{equation}
    \frac{\partial \mathcal{L}}{\partial \bm{A}} = 4 \bm{A} \sum\nolimits_{i=1}^n (f(\bm{x}_i) - y_i) \sum\nolimits_{j=1}^n (f(\bm{x}_j) - y_j) k(\bm{x}_i, \bm{x}_j) (\bm{x}_i - \bm{x}_j) (\bm{x}_i - \bm{x}_j)^T.
\end{equation}

Both KR and reciprocal distance-weighted $k$-NN models produce predictions in the form of
\begin{equation}
   f(\bm{x}_i) = \sum\nolimits_{j=1}^n w(d_{\bm{M}}(\bm{x}_i, \bm{x}_j)) y_j,
\end{equation}
where $w(d_{\bm{M}}(\bm{x}_i, \bm{x}_j))$ is a weighing function of the distance between test and train instances; 
\begin{equation}
  w(d_{\bm{M}}(\bm{x}_i, \bm{x}_j)) = \frac{d_{\bm{M}}^{-1}(\bm{x}_i, \bm{x}_j)}{\sum_{l \in \mathcal{N}_k(\bm{x}_i, d_{\bm{M}})} d_{\bm{M}}^{-1}(\bm{x}_i,  \bm{x}_l) }
\end{equation}
in the case of $k$-NN and 
\begin{equation}
    w(d_{\bm{M}}(\bm{x}_i, \bm{x}_j)) = \frac{k(d_{\bm{M}}(\bm{x}_i, \bm{x}_j))}{\sum_{l=1}^n k(d_{\bm{M}}(\bm{x}_i, \bm{x}_l))}
\end{equation} in the case of kernel regression.
In other terms, with an RBF kernel, the weights of the kernel regression model correspond to a softmax function of the distance.
If we make the (admittedly strong) assumption that the kernel similarities of the structures drop off quickly as the distance increases, we can see how the set of non-zero kernel elements overlaps with the set of nearest neighbours for the $k$-NN model.
Furthermore, while the weighing functions are not identical, they are still monotonically decreasing as $d_{\bm{M}}$ increases.
Hence, we argue that the metric learned by the MLKR metric learning algorithm is also helpful for the $k$-NN.
Indeed, in Section \ref{sec:results}, we demonstrate empirically how the $k$-NN model with MLKR learned metric can, in fact, outperform the corresponding kernel regression model.

Learning the MLKR metric introduces an additional computational cost that scales as $\mathcal{O}(n^2p^2)$, where $p$ is the dimensionality (i.e., the number of features) of the input data.
However, as Weinberger et al. \cite{weinberger2007metric} have shown, $p$ can be limited via the definition of the matrix $\bm{M}$ and with negligible accuracy penalty.
This is conceptually similar to using PCA to reduce the dimensionality of the data.
In our case, we found that limiting $p$ to $50$ did not affect performance; hence, we use this value in all subsequent calculations.

The main hyperparameter in a $k$-NN model is, naturally, the number of neighbours $k$.
Recent work by \citeauthor{kanagawa2024fast} \cite{kanagawa2024fast} has demonstrated how tuning $k$ can be achieved with minimal additional computational cost, leveraging the properties of the $k$-NN formulation.
Indeed, calculating the leave-one-out cross-validation score for a dataset with a $k$-NN model is equivalent to scaling the prediction from a $(k+1)$-NN model.
Hence, if we calculate the full distance matrix between training datapoints once, an $\mathcal{O}(n^2)$ operation, we can then choose an optimal $k$ with negligible computation.
Alternatively, we can use a separate validation set and choose $k$ that provides the smallest validation set loss, which is typically preferable if $n$ is large.

Another benefit of $k$-NN models is that they lend themselves very naturally to uncertainty quantification.
As the prediction of such a model is constructed based on taking the mean of the set of nearest neighbours, we can estimate other statistical properties of the same set, such as variance or quantiles, just as easily.
Admittedly, if the number of nearest neighbours is small, these estimates will be inaccurate.
However, with sufficiently dense and smooth data, higher values of $k$ are preferred, and the problem is less pronounced.
Alternatively, for uncertainty quantification, more neighbours may be included, and the neighbours can be sampled using procedures such as bootstrap \cite{efron1993introduction} or jackknife \cite{efron1982jackknife}.

\subsection{Molecular descriptors}\label{ssec:descriptors}

To employ machine learning on the task of predicting chemical properties, we need a way to express the chemical system numerically.
Many ways to translate the complex three-dimensional structures and physical properties into such representations or \emph{descriptors} exist.
In general, a descriptor is a vector or a tensor which encodes chemical information of the system.
Initially, much emphasis in machine learning for chemistry was placed on \emph{global} descriptors \cite{musil2021physics}, which encode each system using a set of common features, due to their ease of use with standard machine learning approaches.
While global descriptors have been very successful, the so-called local descriptors have received increased attention recently.
Instead of encoding the entire chemical system as a whole, local descriptors instead describe the system through encoding individual atoms or groups thereof.
Local descriptors are naturally better suited for handling datasets with varying sizes of chemical systems, as they inherently express systems as a series of smaller features.
Furthermore, if the target property can be assumed to be decomposable into a sum of atomic contributions, machine learning methods can be adapted to learn these atomic contributions instead of directly inferring the label from the whole system.
This has been shown to work well in practice.
In particular, kernel-based methods such as KRR can easily be adapted to benefit from this additive property \cite{langer2022Representations}.

In this work, we focus mainly on the FCHL representation \cite{faber2018alchemical, christensen2020FCHL} (both local and global variants).
The FCHL representation has previously been tested and found to perform well with KRR, specifically in predicting the properties of atmospheric clusters. \cite{kubecka2022Quantum}
There exist two formulations of the FCHL representation: the original, later renamed FCHL18 by the publication year, and the more efficient, discretized version, termed FCHL19.
In both FCHL representations, the atomic system is expressed as a combination of normal distributions over the first three M-body interatomic expansions of the system.
The first-order expansion encodes the properties of individual atoms, i.e., normal distributions along the rows and columns of the periodic table, whereas the two-body expansion corresponds to the interatomic distances.
Finally, the three-body expansion encodes the angular distribution of the system, along with other structural information.
The key assumption behind the descriptor is that many properties can be approximated as the sum of these many-body terms.
FCHL18 utilizes the first three expansions and produces a three-dimensional tensor as the representation for each system.
FCHL19 aims to reduce the size of the representation by discretizing the distributions and omitting the first-order expansion term, with minimal impact on ML predictive accuracy.
The result is a set of vectors which encode the atomic environment of each atom in a chemical system.
While the FCHL18 representation is always local, summing over the atomic vectors of FCHL19 produces a single, global representation of the system, which can be used in a wide range of machine learning methods.

Several other molecular descriptors---namely the Coulomb Matrix (CM), Bag of Bonds (BoB), and Many-Body Distribution Functionals (MBDF)---are described and employed exclusively in the SI.

\section{Methodology}\label{sec:methods}

\subsection{Technical implementation}
In this section, we give a brief overview of the technical implementation of the calculations, the results of which we present in the following section.

In our calculations, we experimented with different molecular descriptors and found FCHL19 to perform consistently best across our set of datasets.
For a performance comparison between representations, we refer to Section~\ref{supsec:representations} in the Supporting Information.
Hence, all ML models use FCHL19 representation as the input data.
KRR and kernel-induced metric-based $k$-NN models use the local (tensor) FCHL19 representation.
For implementations requiring a global representation (MLKR and Euclidean distance-based methods), we sum over atomic contributions in the FCHL19 tensor to get a vector representing the chemical system.

The computations are conducted using five-fold cross-validation, as we found this to provide sufficiently robust statistics with a reasonable computational cost.
After dividing data into $k_\mathrm{CV}=5$ equally sized parts, one chunk is used as hold-out test data.
The size of the test set remains fixed in each of the computations.
To generate learning curves, the size of training data is varied by subsampling the remaining 80\% of the data.

In each of the computations, we use a KRR model with a standard RBF kernel on local representations as a baseline.
The two main hyperparameters for KRR, the RBF kernel standard deviation $\sigma$ and the ridge penalty $\lambda$ were found via grid search on a training set of $4,000$ items with a test set of $1,000$ items.
As for the $k$-NN implementations, we train three distinct models: one using the kernel-induced distance based on the local FCHL19 representation, another model using the MLKR metric learning algorithm of the global FCHL19 representation, and a standard $k$-NN model on the same global representation and Euclidean distance to find the nearest neighbours.
For all the $k$-NN models, we found that weighting the labels of nearest neighbours by the reciprocal of the distance provided higher accuracy than uniform weights.
The number of neighbours for each dataset was found using a five-fold cross-validated hyperparameter search with a random subsample of $5,000$ items.
As MLKR optimises the metric based on kernel regression loss, we also include a kernel regression model with the MLKR metric in our analysis to study whether the $k$-NN step adds or detracts from the performance of the model.

We also tested $\Delta$-learning (discussed in Section~\ref{ssec:instance_based}) with both KRR and $k$-NN models.
In the following section, whenever $\Delta$-learning is applied, we aim to predict the residual, or difference between labels, of two quantum chemical levels of theory.
The computational methods are specified in the corresponding sections.

The calculations were performed on the Grendel cluster (\url{http://www.cscaa.dk/grendel/}), a computing cluster maintained by the Centre for Scientific Computing Aarhus at the University of Aarhus, using Intel Xeon CPUs at $2.6$--$3.0$ GHz clock speed.

The methodology is implemented as a part of the {\sc JK} software framework \cite{kubecka2019configurational, kubecka23jkcs}, and the code to reproduce the calculations presented here can be found at \url{https://github.com/edahelsinki/JK-kNN/}.

\subsection{Datasets}\label{ssec:datasets}

\paragraph{QM9}
The {\sc qm9} dataset\cite{ramakrishnan2014qm9} of $134,000$ small organic molecules is a widely used, high-quality chemical dataset.
The dataset comprises molecules of H, C, F, N, and O ($\leq 9$ atoms), with 15 properties calculated at the B3LYP/6-31G(2df,p) level of theory.
As this dataset has been well studied in the literature, we include it as a benchmark for our method.
As the target label, we use atomization energy (internal energy at 0 K) directly (i.e., without applying $\Delta$-learning).

\paragraph{Sulphuric acid--water systems}
Sulphuric acid (SA) is known to be a main contributor to new particle formation \cite{paasonen2010roles} (NPF), both in continental regions and over oceans.
It has a low saturation vapour pressure and a high ability to form molecular clusters with many other molecules, such as various bases.
The sulphuric acid--water (SA--W) system is considered the simplest system relevant to understanding the first steps of NPF.
Even for such a simple binary system, there are numerous possible configurations of the molecular clusters, which makes KRR methods computationally unfeasible.

We reused the database constructed by Kube{\v c}ka et al.\cite{Kubecka_estlett_2022}, where representative SA--W cluster configurations are evaluated at $\omega$B97X-D/6-31++G(d,p) and GFN1-xTB levels of theory.
The former is a commonly applied and well-benchmarked density functional theory (DFT) method suitable for sizeable atmospheric molecular cluster systems.\cite{Elm2013,Elm2017,schmitz2020,Jensen2022}
As the basis set is relatively small, the energy is often refined at a higher level of theory, but this method is known to provide accurate equilibrium geometries. 
The latter is less accurate but a fast semi-empirical method (using a tight-binding DFT approach). 

We examined ML techniques either by directly predicting the electronic binding energy of the SA--W clusters at the $\omega$B97X-D/6-31++G(d,p) level of theory or by using a $\Delta$-learning approach, where the residual is between the aforementioned level of theory and the lower GFN1-xTB level.

\paragraph{Clusterome}
The main benefit of the increased computational efficiency of a machine learning model is that it allows the model to handle larger datasets.
Therefore, the final dataset we examine, \emph{Clusterome}, was chosen to demonstrate the scalability of our approach.
Clusterome is an atmospheric cluster dataset produced using the Clusteromics I--V datasets \cite{Clusteromics_I, Clusteromics_II, Clusteromics_III, Clusteromics_IV, Clusteromics_V} and published in a combined form by Knattrup et al.\cite{knattrup2023clusterome}.
This dataset consists of unique atmospheric acid--base cluster structures, with sulphuric acid (SA), methanesulphonic acid (MSA), nitric acid (NA), and formic acid (FA) as acids and ammonia (AM), methylamine (MA), dimethylamine (DMA), trimethylamine (TMA), and ethylenediamine as bases. 
The Clusteromics I--V datasets contain $22,870$ equilibrium structures obtained at the $\omega$B97X-D/6-31++G(d,p) level of theory, and Knattrup et al. expanded this dataset to $251,554$ entries by including out-of-equilibrium structures obtained from short MD simulations performed at the GFN1-xTB level of theory.
In the calculations, we use the $\Delta$-learning approach with the target being the residual between the $\omega$B97X-D/6-31++G(d,p) (high) and GFN1-xTB (low) levels of theory.

\section{Results and Discussion}
\label{sec:results}

\subsection{Learning atmospheric molecular clusters}

\subsubsection{Accuracy and computational cost}
\label{ssec:saw-results}

\begin{figure}
    \centering
    \includegraphics[width=0.85\linewidth]{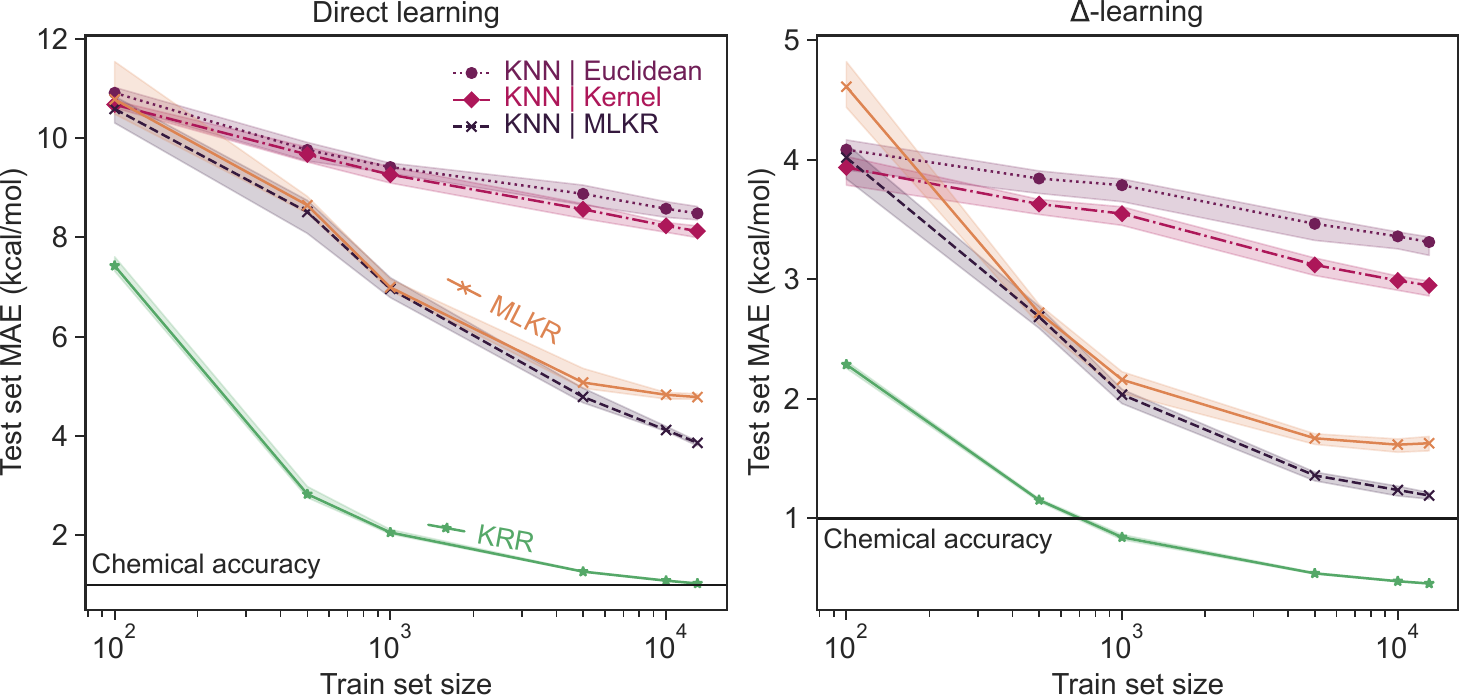}
    \caption{Learning curves, or mean absolute error (MAE) as a function of training set size for direct- and $\Delta$-learning of electronic binding energies for sulphuric acid--water clusters. The black solid line denotes chemical accuracy ($\textrm{MAE} = 1$~kcal/mol).}
    \label{fig:lr_saw_delta}
\end{figure}

First, we compare the performance of different ML models with both direct and $\Delta$-learning on the dataset of sulphuric acid--water (SA--W) clusters.
Fig.~\ref{fig:lr_saw_delta} shows the learning curves for the KRR, MLKR, and $k$-NN methods.
$\Delta$-learning leads to a shift in the learning curves with $\sim$2-times better accuracy compared to direct-learning.
However, the overall trends remain the same.
Similar to Kube{\v c}ka et al. \cite{kubecka2022Quantum}, the KRR model nearly achieves chemical accuracy with $n=13,000$ training items when using the labels from the high level of theory directly, with mean absolute error decreasing to $0.46$~kcal/mol when using $\Delta$-learning.

As for the $k$-NN implementations, the approaches based on MLKR metric learning outperform both the kernel-induced distance and standard $k$-NN variants.
In direct learning, the MLKR-based $k$-NN model reaches a mean absolute error of $3.86$~kcal/mol.
In $\Delta$-learning, the difference in performance between KRR and $k$-NN models is less pronounced, with the best $k$-NN implementation (MLKR with FCHL19) nearly reaching the mark for chemical accuracy of mean absolute error of 1~kcal/mol.
Additionally, pairing the $k$-NN model to the MLKR outperforms the MLKR-based kernel regression model in both direct and $\Delta$-learning, with the gap growing wider as the size of training data increases.

\begin{figure}
    \centering
    \includegraphics[width=0.85\linewidth]{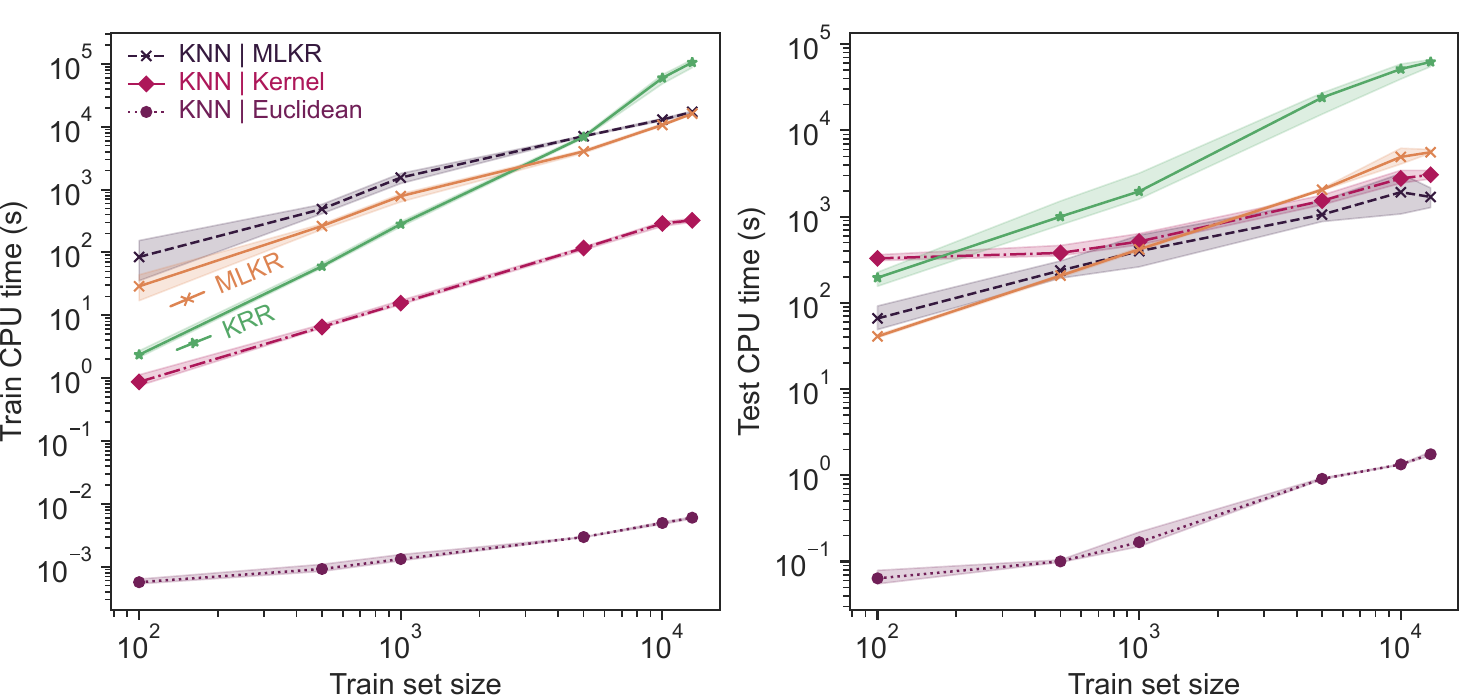}
    \caption{Computation times for direct-learning of electronic binding energies for SA--W clusters. For simplicity, we present computation times only for direct-learning, as direct- and $\Delta$-learning times are very similar. Notice the logarithmic scale on the y-axis.}
    \label{fig:lr_saw_time}
\end{figure}

While the $k$-NN models do not reach the same level of accuracy as KRR models, as presented in Fig.~\ref{fig:lr_saw_time}, $k$-NN may have a significant computational edge on large databases.
Naturally, the Euclidean distance-based $k$-NN far overperforms all other variants in computational costs.
However, as predicted by the theory, the training time for KRR and the other $k$-NN approaches scales differently, with KRR requiring 100 times more CPU time than the slowest $k$-NN implementation already $n=5,000$ training data points.
Moreover, as the KRR training in the limit scales as ${\mathcal O}(n^3)$ with respect to the dataset size, the method quickly becomes entirely computationally intractable.
For the learning algorithms paired with MLKR, there is a decrease of two orders of magnitude in inference time compared to KRR.
More crucially, the inference time also shows similar speed gain for the MLKR metric learning algorithms, with a smaller yet substantial increase for the kernel-induced distance approach as well.
Overall, based on the results presented, the MLKR-based $k$-NN approach combines good predictive performance with impressive computational efficiency.

The kernel-induced distance-based $k$-NN model does not significantly outperform the Euclidean-distance-based variant.
This was unexpected given the strong performance of the KRR model using the kernel as a similarity metric.
We offer two hypotheses for the poor performance.
First, the KRR model can take into account the entire training data, whereas a $k$-NN necessarily has a hard cut-off and cannot use information beyond the $k$ nearest neighbours.
Additionally, the kernel-induced distance may not align with kernel values in unnormalised extensive kernels.
We also experimented with normalising the kernel before calculating the kernel-induced distance, but this did not improve results.
Normalizing the kernel did not improve results, suggesting useful information for the $k$-NN model is lost.

\subsubsection{Number of nearest neighbours}
\begin{figure}[htbp]
    \centering
    \includegraphics[width=0.99\linewidth]{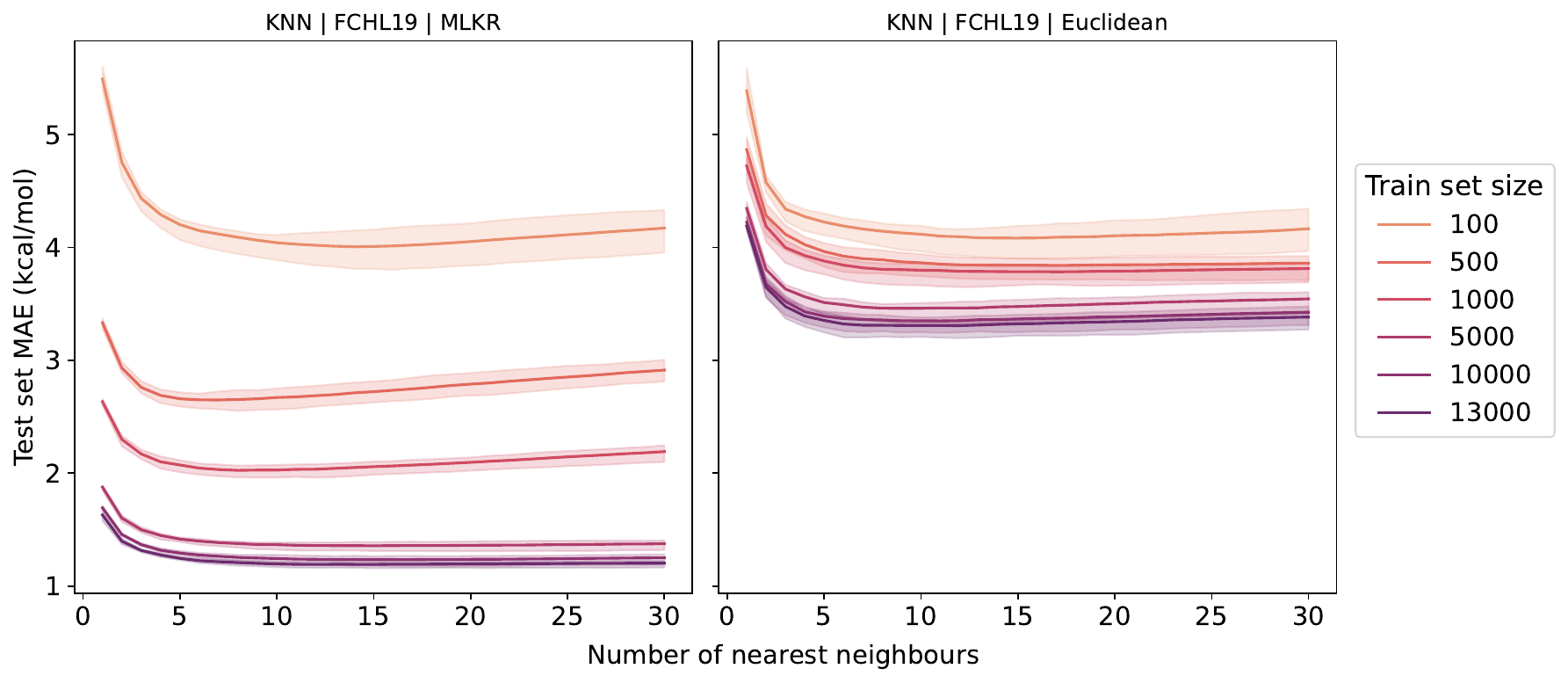}
    \caption{Test mean absolute error as a function of $k$ at different training set sizes (coloured lines) on the SA--W $\Delta$-learning data for both the MLKR-based $k$-NN and the Euclidean variant. The choice of $k$ has little impact, especially as the dataset size grows, provided it is reasonably chosen (i.e., $5 \leq k \leq 15$).} 
    \label{fig:k-scaling}
\end{figure}

The choice of how many nearest neighbours to consider ($k$) is pivotal for the model's performance. 
A small $k$ represents a more flexible model, but is more prone to overfitting; a high value for $k$ results in more robust but less expressive models.
In the learning curve calculations above, the value of $k=12$ was chosen based on a hyperparameter search described in Section \ref{sec:methods}.
To ensure that the optimal $k$ does not depend on the size of the training dataset, we re-ran the calculations for different values of $k$.
Fig.~\ref{fig:k-scaling} shows how the mean absolute error of various $k$-NN implementations rapidly decreases as $k$ increases but quickly plateaus for the MLKR-based $k$-NN model.
Furthermore, the optimal $k$ value remains practically constant across different dataset sizes.
This suggests that, especially for large datasets, the model performance is not sensitive to the choice of $k$, provided that it is reasonable (e.g., $5 \leq k \leq 15$).
Hence, for users, if hyperparameter optimisation (e.g., using a similar procedure as described here) is infeasible, we recommend setting $k=10$ as the first guess when using an MLKR-based $k$-NN model.

The Euclidean variant is more sensitive to the choice of $k$, even at $13,000$ items.
Nonetheless, the optimal value of $k$ remains constant even for this variant.

\subsubsection{Modelling large clusters}

\begin{figure}[htbp]
    \centering
    \includegraphics[width=0.5\linewidth]{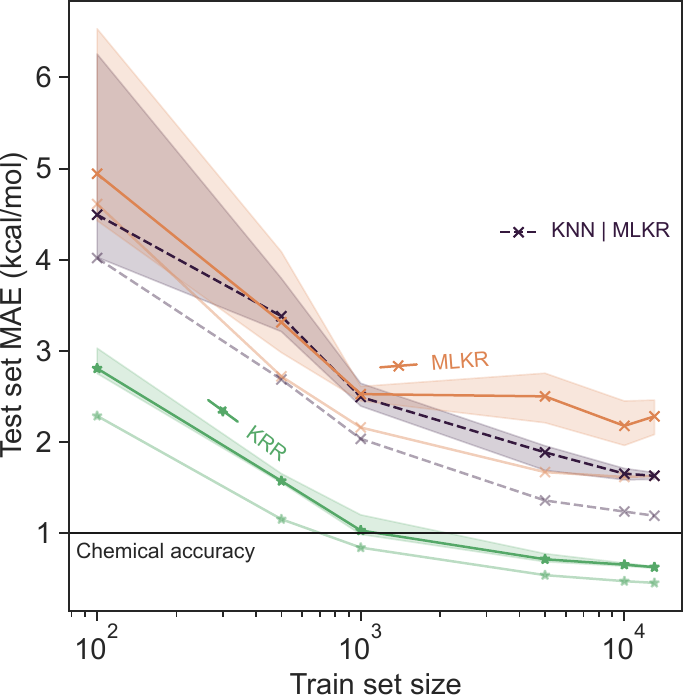}
    \caption{The extrapolation - learning curves for the SA--W cluster, where instead of using cross-validation (cf. Fig.~\ref{fig:lr_saw_delta}), the largest (SA)$_4$(W)$_5$ clusters are used for test, while the remaining smaller clusters are used for training. The faint lines show the learning curves for $\Delta$-learning on the cross-validated data.}
    \label{fig:saw_extrapolation}
\end{figure}

To compare the extrapolation performance of various $k$-NN implementations against KRR, we trained models on the SA--W data with $\Delta$-learning similar to Section~\ref{ssec:saw-results}, excluding the largest (SA)$_4$(W)$_5$ clusters, and then attempted to predict the electronic energies of these holdout datapoints.
As Fig.~\ref{fig:saw_extrapolation} shows, the difference between KRR and $k$-NN is more pronounced in extrapolation compared to interpolation (cf. Fig.~\ref{fig:lr_saw_delta}); the KRR model reaches a mean absolute error of $0.63$ kcal/mol at $13,000$ training items while the best-performing MLKR-based $k$-NN model has an error of $1.63$ kcal/mol at the same level.
Nonetheless, the $k$-NN model maintains acceptable performance even for extrapolation.
The relative accuracy drop between the extrapolation task and the easier cross-validation task (shown as faint lines in Fig.~\ref{fig:saw_extrapolation}) is similar for both KRR and $k$-NN models.

\subsubsection{Interpretability and uncertainty estimation}

The nearest neighbour sets of a $k$-NN model can be analysed for model interpretation and uncertainty estimation.
From these sets, for example, we may estimate the deciles as a representation of the prediction uncertainty.
Furthermore, the set of nearest neighbours can be manually inspected to learn which items the model deems most similar and to ensure that this neighbour set aligns with domain expectations.

\begin{figure}
    \centering
    \includegraphics[width=0.99\linewidth]{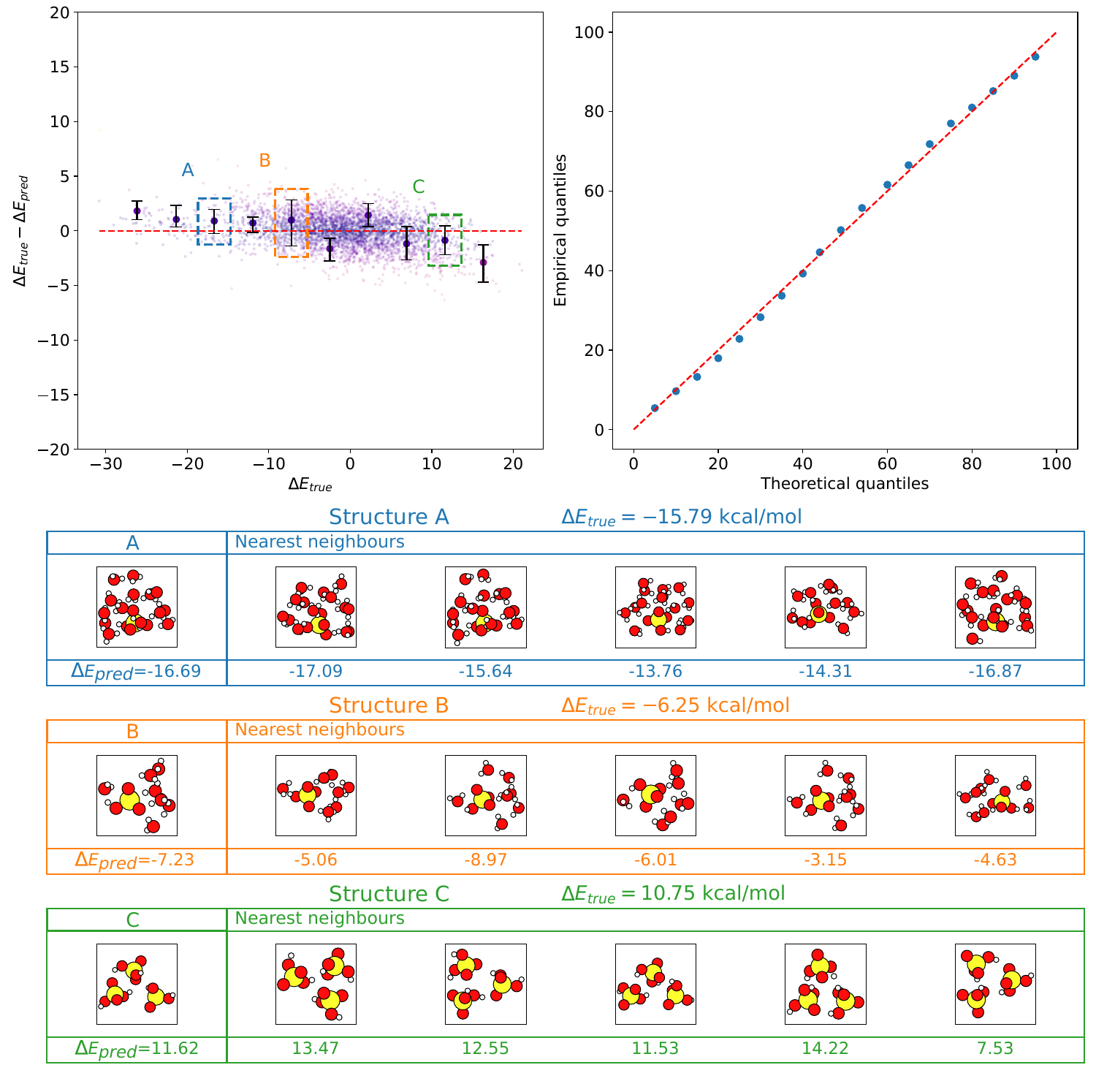}
    \caption{Demonstration of uncertainty estimation and interpretations of a $k$-NN model. The upper left panel shows the true label value vs. absolute error of an MLKR-based $k$-NN model trained on the SA--W data, with ten items additionally showing error bars based on 50\% confidence intervals from the set of nearest neighbours for each item, and a red line indicating zero error. The upper right panel shows a calibration plot for the quantiles, where blue dots show the percentage of test labels falling below each empirical percentile. Finally, the bottom half shows the structures, true, and predicted label values for three example items in the test set (indicated by coloured boxes).}
    \label{fig:interpretability_uncertainty}
\end{figure}

To demonstrate these qualities, we study a cross-validation fold from the $\Delta$-learning calculation using the SA--W dataset, trained on $13,000$ structures and tested on $3,471$ previously unseen structures.
The $k$-NN model in question is MLKR-based, and we used the FCHL representation.
In the upper left corner of Fig.~\ref{fig:interpretability_uncertainty}, we show a scatter plot of the true label value of items in the test set vs. the residual (error) between the true value and the prediction.
Ten items show the 50\% confidence intervals, i.e., the range between the 25th and 75th percentiles of the labels in the neighbour set of these ten items, as error bars, with a red dashed line indicating zero error.
For most items, the true label is either included in or close to the error bars, implying that the confidence intervals derived from nearest neighbours are meaningful.

To further ensure that the quantiles for the target value from the neighbour sets are properly calibrated, the upper right corner in the same figure shows a calibration plot for the quantiles.
In this context, calibrated means that our quantiles correspond to the statistical properties of the data.
For each item, we first estimate quantiles based on the set of unweighted nearest neighbours in the training dataset.
We then compare these quantiles to the true labels on the test set, which are previously unseen by the model.
If the true label is below the estimated 5th percentile approximately in one in twenty items (i.e., 5\%) in the test set and so on for other percentiles, the quantiles can be considered well calibrated.
As the upper right panel shows, the proportion of items in the test below the corresponding percentile value (blue dots) follow the theoretical percentiles (red dashed line) closely, demonstrating good calibration.

Finally, the bottom half of Fig.~\ref{fig:interpretability_uncertainty} shows the five nearest neighbours for three example structures, indicated by coloured boxes in the upper left panel.
As can be seen, the model has learned to find structures of similar composition.
We will not delve further into analysing the chemical properties of these particular structures as that is not the main aim of the paper; we seek to demonstrate how the $k$-NN modelling approach allows for interpretation of results by examination of the nearest neighbour set by the user.

\subsection{Learning large datasets}

\begin{figure}
    \centering
    \includegraphics[width=0.99\linewidth]{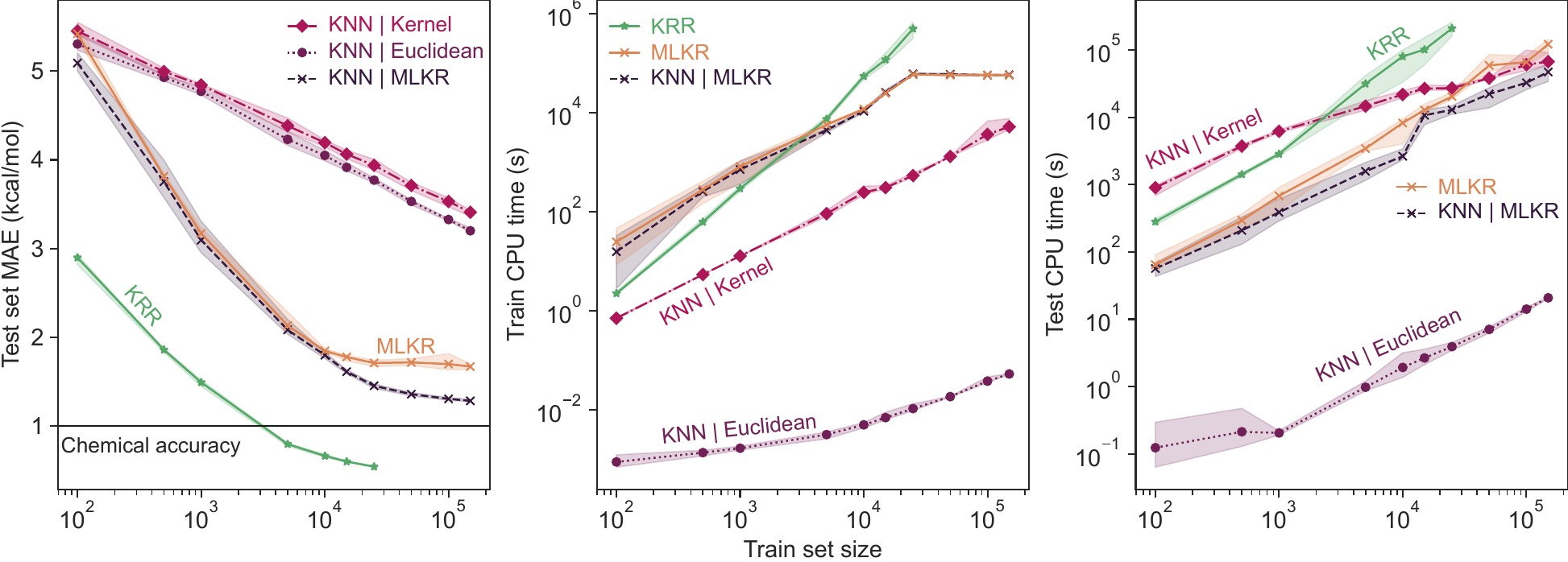}
    \caption{Learning curves and computational cost for learning the electronic binding energies of the Clusterome dataset. The computational advantage of $k$-NN approaches can be seen clearly. In MLKR and MLKR-based $k$-NN we have limited the metric learning to a subsample of $25,000$ items.}
    \label{fig:lr_clusteromicsiv}
\end{figure}

The Clusterome dataset is both a larger and more complex version of the SA--W cluster data discussed in the previous section. Figure \ref{fig:lr_clusteromicsiv} presents the learning curve and computational cost for the tested models on the Clusterome dataset.
Expanding the training set size beyond $100,000$ clusters increases the computational cost, which should give the $k$-NN models an advantage.
Furthermore, we found that capping the MLKR metric learning to a random subsample of $25,000$ items introduces only a negligible cost to accuracy; the $k$-NN model then uses the learned metric with all available training data.
For example, at $50,000$ training items, using the full training set to learn the MLKR metric resulted in a MAE of $1.30$~kcal/mol in contrast to $1.36$~kcal/mol for the subsampled variant; only a $4.6$\% decrease.
Performing subsampling does, however, cut the training time drastically; as expected, for the set $50,000$ items, the subsampled variant required, on average, only $26$\% of the time needed to train the non-subsampled variant.

The larger training set size illustrates the difference in scaling computational cost.
As the size of the training set grows, the difference in training time between MLKR-based $k$-NN models and KRR widens by orders of magnitude, even before considering the subsampling for MLKR metric learning.
Accuracy-wise, while the KRR model outperforms the $k$-NN approaches here, as well, we can still nearly reach chemical accuracy with the MLKR-based $k$-NN model while handling many times more data than would be computationally feasible for KRR.

\subsection{Learning the QM9 compound dataset}

\begin{figure}[htbp]
    \centering
    \includegraphics[width=0.5\linewidth]{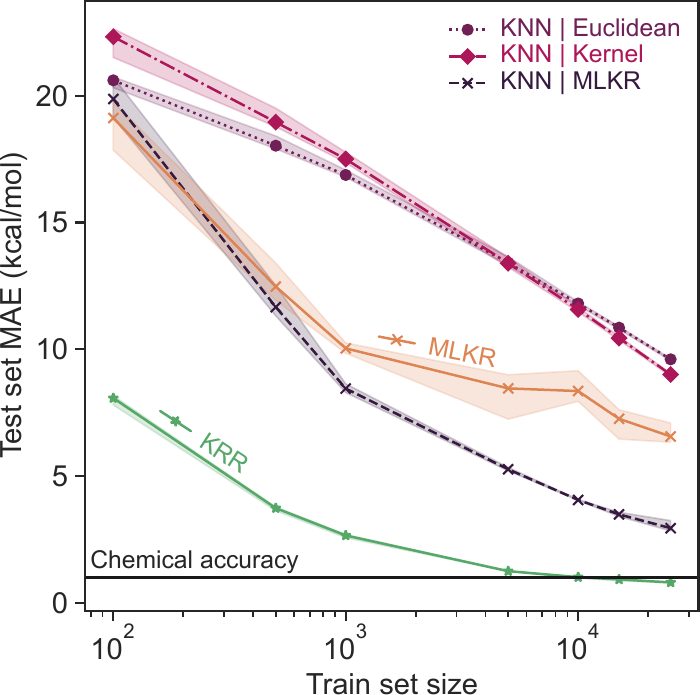}
    \caption{Learning curves for the atomization energies of the {\sc qm9} dataset. The shaded area reflects the error variance.}
    \label{fig:lr_qm9}
\end{figure}

The learning curves for all models trained on the {\sc qm9} dataset are shown in Fig.~\ref{fig:lr_qm9}.
As mentioned in Section~\ref{ssec:datasets}, the test set size is fixed to $26,777$ items; the training set size varies.
The KRR model shows similar performance to Christensen et al.\cite{christensen2020FCHL}, who introduced the FCHL19 representation, ensuring that our implementation is correct.
The KRR model reaches the level of MAE $\approx 1$~kcal/mol with around $10,000$ training samples.
The best-performing $k$-NN variant, using MLKR metric learning, reaches an MAE of $3$~kcal/mol only at $25,000$ samples, albeit with much lower computational cost.
Training for both MLKR-based $k$-NN model and the KRR model at $25,000$ training samples took $60,000$ CPU-seconds, on average, and predicting the test set of $26,775$ points took $57,000$ CPU-seconds for the $k$-NN and $162,000$ CPU-seconds for the KRR; nearly triple the computational cost.

As expected, the Euclidean-distance-based $k$-NN model performs far worse than the metric-learning variant, with a top mean absolute error of $9.6$~kcal/mol, albeit with minimal computational cost; training the model at $25,000$ training items took less than $0.01$ CPU-seconds and predicting the test set less than $20$ CPU-seconds, on average.
The kernel-induced distance-based $k$-NN model shows similar performance to the Euclidean-distance-based variant.

Finally, as before, the results show how the MLKR-based kernel regression model does not reach the accuracy of the MLKR-equipped $k$-NN model, demonstrating that combining the models yields gains in accuracy.
As most of the computational budget is spent on learning the distance metric, the computational cost of MLKR-kernel regression is very similar to the $k$-NN variant both in training and prediction.

\subsection{Discussion and outlook}

Finding novel, efficient, and interpretable ways to predict the properties of large chemical systems would unlock greater understanding of the complex processes affecting the climate and our health.
In this section, we have demonstrated how $k$-NN regression, a classic yet often overlooked machine learning approach, offers a highly efficient and more interpretable alternative to the widely used KRR models as a tool for machine learning in chemistry.
By integrating chemically informed distance metrics---in particular MLKR-based metric learning---we show that \textit{k}-NN models can achieve near-chemical accuracy while dramatically reducing computational overhead. 
In particular, our approach scales effortlessly to datasets exceeding 250,000 entries and performs robustly even in extrapolation tasks involving large molecular clusters.
This enhanced computational efficiency, combined with model transparency, makes such models an enticing alternative, especially in data-rich, interpolation-oriented settings.

As curated datasets continue to grow, we anticipate that hybrid workflows leveraging fast, interpretable instance-based models like \textit{k}-NN will become an increasingly valuable component of molecular computational chemistry pipelines with further applications, e.g., in atmospheric chemistry.
Future work could look into ways to encode prior information (such as cluster composition) more explicitly into the metric learning layer or coupling it with scalable approximate nearest-neighbour methods for ultra-large-scale applications.
Notably, applying \textit{k}-NN-based models to molecular dynamics (MD) simulations---by learning not only energies but also forces---would be a promising avenue.
This would allow for transparent and computationally efficient MD models without relying on less interpretable architectures, such as neural networks, and settings where KRR becomes computationally intractable.

\begin{acknowledgement}
This work was funded by the Research Council of Finland for the VILMA Centre for Excellence (project 364226) and Danish National Research Foundation (DNRF172) through the Center of Excellence for Chemistry of Clouds.%

The numerical results presented in this work were obtained at the Centre for Scientific Computing, Aarhus (\url{https://phys.au.dk/forskning/faciliteter/cscaa/}).
 
\end{acknowledgement}

\begin{suppinfo}
The following is available as supporting information:
\begin{itemize}    
    \item All the code necessary to recreate the calculations is available at \url{https://github.com/edahelsinki/JK-kNN/}.
    \item A comparison between the performance of molecular representations other than FCHL19 can be found in Section \ref{supsec:representations}.
\end{itemize}

\end{suppinfo}

\large
\noindent\textbf{Author Information}
\\
\normalsize
\textbf{Corresponding Author}
\\
Lauri Seppäläinen - Department of Computer Science, Pietari Kalmin katu 5, University of Helsinki, Helsinki, FI 00560; Email: lauri.seppalainen@helsinki.fi
\\
\normalsize
\textbf{Authors}
\\
Jakub Kube{\v c}ka - Department of Chemistry, Aarhus University, Langelandsgade 140, Aarhus C, DK 8000; Email: ja-kub-ecka@chem.au.dk\\
Jonas Elm - Department of Chemistry, Aarhus University, Langelandsgade 140, Aarhus C, DK 8000; Email: jelm@chem.au.dk\\
Kai R. Puolamäki - Department of Computer Science, Pietari Kalmin katu 5, University of Helsinki, Helsinki, FI 00560; Email: kai.puolamaki@helsinki.fi

\bibliography{clusters}

\providecommand{\latin}[1]{#1}
\makeatletter
\providecommand{\doi}
  {\begingroup\let\do\@makeother\dospecials
  \catcode`\{=1 \catcode`\}=2 \doi@aux}
\providecommand{\doi@aux}[1]{\endgroup\texttt{#1}}
\makeatother
\providecommand*\mcitethebibliography{\thebibliography}
\csname @ifundefined\endcsname{endmcitethebibliography}  {\let\endmcitethebibliography\endthebibliography}{}
\begin{mcitethebibliography}{75}
\providecommand*\natexlab[1]{#1}
\providecommand*\mciteSetBstSublistMode[1]{}
\providecommand*\mciteSetBstMaxWidthForm[2]{}
\providecommand*\mciteBstWouldAddEndPuncttrue
  {\def\EndOfBibitem{\unskip.}}
\providecommand*\mciteBstWouldAddEndPunctfalse
  {\let\EndOfBibitem\relax}
\providecommand*\mciteSetBstMidEndSepPunct[3]{}
\providecommand*\mciteSetBstSublistLabelBeginEnd[3]{}
\providecommand*\EndOfBibitem{}
\mciteSetBstSublistMode{f}
\mciteSetBstMaxWidthForm{subitem}{(\alph{mcitesubitemcount})}
\mciteSetBstSublistLabelBeginEnd
  {\mcitemaxwidthsubitemform\space}
  {\relax}
  {\relax}

\bibitem[Kulmala(2025)]{kulmala25}
Kulmala,~M. Importance of new particle formation for climate and air quality. \emph{ACS ES\&T Air} \textbf{2025}, \emph{2}, 710--712\relax
\mciteBstWouldAddEndPuncttrue
\mciteSetBstMidEndSepPunct{\mcitedefaultmidpunct}
{\mcitedefaultendpunct}{\mcitedefaultseppunct}\relax
\EndOfBibitem
\bibitem[Zhao \latin{et~al.}()Zhao, Donahue, Zhang, Mao, Shrivastava, Ma, Shen, Wang, Sun, Gordon, Tang, Fast, Wang, Gao, Yan, Singh, Li, Huang, Lou, Lin, Wang, Jiang, Ding, Nie, Qi, Chi, and Wang]{zhao2024global}
Zhao,~B.; Donahue,~N.~M.; Zhang,~K.; Mao,~L.; Shrivastava,~M.; Ma,~P.-L.; Shen,~J.; Wang,~S.; Sun,~J.; Gordon,~H. \latin{et~al.}  Global Variability in Atmospheric New Particle Formation Mechanisms. \emph{631}, 98--105\relax
\mciteBstWouldAddEndPuncttrue
\mciteSetBstMidEndSepPunct{\mcitedefaultmidpunct}
{\mcitedefaultendpunct}{\mcitedefaultseppunct}\relax
\EndOfBibitem
\bibitem[Cai \latin{et~al.}(2024)Cai, Sulo, Gu, Holm, Cai, Thomas, Neuberger, Mattsson, Paglione, Decesari, Rinaldi, Yin, Aliaga, Huang, Li, Gramlich, Ciarelli, Qu\'el\'ever, Sarnela, Lehtipalo, Zannoni, Wu, Nie, Kangasluoma, Mohr, Kulmala, Zha, Stolzenburg, and Bianchi]{cai24}
Cai,~J.; Sulo,~J.; Gu,~Y.; Holm,~S.; Cai,~R.; Thomas,~S.; Neuberger,~A.; Mattsson,~F.; Paglione,~M.; Decesari,~S. \latin{et~al.}  Elucidating the mechanisms of atmospheric new particle formation in the highly polluted Po Valley, Italy. \emph{Atmos. Chem. Phys.} \textbf{2024}, \emph{24}, 2423--2441\relax
\mciteBstWouldAddEndPuncttrue
\mciteSetBstMidEndSepPunct{\mcitedefaultmidpunct}
{\mcitedefaultendpunct}{\mcitedefaultseppunct}\relax
\EndOfBibitem
\bibitem[Merikanto \latin{et~al.}(2009)Merikanto, Spracklen, Mann, Pickering, and Carslaw]{CCN_budget}
Merikanto,~J.; Spracklen,~D.~V.; Mann,~G.~W.; Pickering,~S.~J.; Carslaw,~K.~S. Impact of nucleation on global {CCN}. \emph{Atmos. Chem. Phys.} \textbf{2009}, \emph{9}, 8601--8616\relax
\mciteBstWouldAddEndPuncttrue
\mciteSetBstMidEndSepPunct{\mcitedefaultmidpunct}
{\mcitedefaultendpunct}{\mcitedefaultseppunct}\relax
\EndOfBibitem
\bibitem[Lohmann and Feichter(2005)Lohmann, and Feichter]{CCN}
Lohmann,~U.; Feichter,~J. Global indirect aerosol effects: A review. \emph{Atmos. Phys. Chem.} \textbf{2005}, \emph{5}, 715--737\relax
\mciteBstWouldAddEndPuncttrue
\mciteSetBstMidEndSepPunct{\mcitedefaultmidpunct}
{\mcitedefaultendpunct}{\mcitedefaultseppunct}\relax
\EndOfBibitem
\bibitem[Tr{\"o}stl \latin{et~al.}(2016)Tr{\"o}stl, Chuang, Gordon, Heinritzi, Yan, Molteni, Ahlm, Frege, Bianchi, Wagner, and {et al.}]{Trostl_nature_2016}
Tr{\"o}stl,~J.; Chuang,~W.~K.; Gordon,~H.; Heinritzi,~M.; Yan,~C.; Molteni,~U.; Ahlm,~L.; Frege,~C.; Bianchi,~F.; Wagner,~R. \latin{et~al.}  The Role of Low-volatility Organic Compounds in Initial Particle Growth in the Atmosphere. \emph{Nature} \textbf{2016}, \emph{533}, 527--531\relax
\mciteBstWouldAddEndPuncttrue
\mciteSetBstMidEndSepPunct{\mcitedefaultmidpunct}
{\mcitedefaultendpunct}{\mcitedefaultseppunct}\relax
\EndOfBibitem
\bibitem[Canadell \latin{et~al.}(2021)Canadell, Monteiro, Costa, Cotrim~da Cunha, Cox, Eliseev, Henson, Ishii, Jaccard, Koven, and {et al.}]{IPCC2021new}
Canadell,~J.~G.; Monteiro,~P. M.~S.; Costa,~M.~H.; Cotrim~da Cunha,~L.; Cox,~P.; Eliseev,~A.~V.; Henson,~S.; Ishii,~M.; Jaccard,~S.; Koven,~C. \latin{et~al.}  In \emph{Climate Change 2021: The Physical Science Basis. Contribution of Working Group I to the Sixth Assessment Report of the Intergovernmental Panel on Climate Change}; Masson-Delmotte,~V., Zhai,~P., Pirani,~A., Connors,~S.~L., Péan,~C., Berger,~S., Caud,~N., Chen,~Y., Goldfarb,~L., Gomis,~M.~I. \latin{et~al.} , Eds.; Cambridge University Press: Cambridge, United Kingdom and New York, NY, USA, 2021; pp 673--816\relax
\mciteBstWouldAddEndPuncttrue
\mciteSetBstMidEndSepPunct{\mcitedefaultmidpunct}
{\mcitedefaultendpunct}{\mcitedefaultseppunct}\relax
\EndOfBibitem
\bibitem[Kulmala \latin{et~al.}()Kulmala, Kontkanen, Junninen, Lehtipalo, Manninen, Nieminen, Petäjä, Sipilä, Schobesberger, Rantala, Franchin, Jokinen, Järvinen, Äijälä, Kangasluoma, Hakala, Aalto, Paasonen, Mikkilä, Vanhanen, Aalto, Hakola, Makkonen, Ruuskanen, Mauldin, Duplissy, Vehkamäki, Bäck, Kortelainen, Riipinen, Kurtén, Johnston, Smith, Ehn, Mentel, Lehtinen, Laaksonen, Kerminen, and Worsnop]{kulmala2013direct}
Kulmala,~M.; Kontkanen,~J.; Junninen,~H.; Lehtipalo,~K.; Manninen,~H.~E.; Nieminen,~T.; Petäjä,~T.; Sipilä,~M.; Schobesberger,~S.; Rantala,~P. \latin{et~al.}  Direct Observations of Atmospheric Aerosol Nucleation. \emph{339}, 943--946\relax
\mciteBstWouldAddEndPuncttrue
\mciteSetBstMidEndSepPunct{\mcitedefaultmidpunct}
{\mcitedefaultendpunct}{\mcitedefaultseppunct}\relax
\EndOfBibitem
\bibitem[Lee \latin{et~al.}(2019)Lee, Gordon, Yu, Lehtipalo, Haley, Li, and Zhang]{lee19}
Lee,~S.-H.; Gordon,~H.; Yu,~H.; Lehtipalo,~K.; Haley,~R.; Li,~Y.; Zhang,~R. New particle formation in the atmosphere: From molecular clusters to global climate. \emph{J. Geophys. Res. Atmos} \textbf{2019}, \emph{124}, 7098--7146\relax
\mciteBstWouldAddEndPuncttrue
\mciteSetBstMidEndSepPunct{\mcitedefaultmidpunct}
{\mcitedefaultendpunct}{\mcitedefaultseppunct}\relax
\EndOfBibitem
\bibitem[Riplinger and Neese(2013)Riplinger, and Neese]{riplinger13a}
Riplinger,~C.; Neese,~F. An efficient and near linear scaling pair natural orbital based local coupled cluster method. \emph{J. Chem. Phys.} \textbf{2013}, \emph{138}, 034106\relax
\mciteBstWouldAddEndPuncttrue
\mciteSetBstMidEndSepPunct{\mcitedefaultmidpunct}
{\mcitedefaultendpunct}{\mcitedefaultseppunct}\relax
\EndOfBibitem
\bibitem[Riplinger \latin{et~al.}(2013)Riplinger, Sandhoefer, Hansen, and Neese]{riplinger13b}
Riplinger,~C.; Sandhoefer,~B.; Hansen,~A.; Neese,~F. Natural triple excitations in local coupled cluster calculations with pair natural orbitals. \emph{J. Chem. Phys.} \textbf{2013}, \emph{139}, 134101\relax
\mciteBstWouldAddEndPuncttrue
\mciteSetBstMidEndSepPunct{\mcitedefaultmidpunct}
{\mcitedefaultendpunct}{\mcitedefaultseppunct}\relax
\EndOfBibitem
\bibitem[Riplinger \latin{et~al.}(2016)Riplinger, Pinski, Becker, Valeev, and Neese]{riplinger16}
Riplinger,~C.; Pinski,~P.; Becker,~U.; Valeev,~E.~F.; Neese,~F. Sparse maps -- {A} systematic infrastructure for reduced-scaling electronic structure methods. {II}. {L}inear scaling domain based pair natural orbital coupled cluster theory. \emph{J. Chem. Phys.} \textbf{2016}, \emph{144}, 024109\relax
\mciteBstWouldAddEndPuncttrue
\mciteSetBstMidEndSepPunct{\mcitedefaultmidpunct}
{\mcitedefaultendpunct}{\mcitedefaultseppunct}\relax
\EndOfBibitem
\bibitem[Tew(2019)]{tew19}
Tew,~D.~P. Principal Domains in Local Correlation Theory. \emph{Journal of Chemical Theory and Computation} \textbf{2019}, \emph{15}, 6597--6606, PMID: 31622093\relax
\mciteBstWouldAddEndPuncttrue
\mciteSetBstMidEndSepPunct{\mcitedefaultmidpunct}
{\mcitedefaultendpunct}{\mcitedefaultseppunct}\relax
\EndOfBibitem
\bibitem[G and C.(2016)G, and C.]{schmitz18}
G,~S.; C.,~H. Perturbative triples correction for local pair natural orbital based explicitly correlated CCSD(F12*) using Laplace transformation techniques. \emph{J Chem Phys.} \textbf{2016}, \emph{145}, 234107\relax
\mciteBstWouldAddEndPuncttrue
\mciteSetBstMidEndSepPunct{\mcitedefaultmidpunct}
{\mcitedefaultendpunct}{\mcitedefaultseppunct}\relax
\EndOfBibitem
\bibitem[Nagy \latin{et~al.}(2018)Nagy, Samu, and Kállay]{nagy2018}
Nagy,~P.~R.; Samu,~G.; Kállay,~M. Optimization of the Linear-Scaling Local Natural Orbital CCSD(T) Method: Improved Algorithm and Benchmark Applications. \emph{Journal of Chemical Theory and Computation} \textbf{2018}, \emph{14}, 4193--4215\relax
\mciteBstWouldAddEndPuncttrue
\mciteSetBstMidEndSepPunct{\mcitedefaultmidpunct}
{\mcitedefaultendpunct}{\mcitedefaultseppunct}\relax
\EndOfBibitem
\bibitem[Nagy and Kállay(2019)Nagy, and Kállay]{nagy19}
Nagy,~P.~R.; Kállay,~M. Approaching the Basis Set Limit of CCSD(T) Energies for Large Molecules with Local Natural Orbital Coupled-Cluster Methods. \emph{Journal of Chemical Theory and Computation} \textbf{2019}, \emph{15}, 5275--5298\relax
\mciteBstWouldAddEndPuncttrue
\mciteSetBstMidEndSepPunct{\mcitedefaultmidpunct}
{\mcitedefaultendpunct}{\mcitedefaultseppunct}\relax
\EndOfBibitem
\bibitem[Kállay \latin{et~al.}(2020)Kállay, Nagy, Mester, Rolik, Samu, Csontos, Csóka, Szabó, Gyevi-Nagy, Hégely, Ladjánszki, Szegedy, Ladóczki, Petrov, Farkas, Mezei, and Ganyecz]{kallay2020}
Kállay,~M.; Nagy,~P.~R.; Mester,~D.; Rolik,~Z.; Samu,~G.; Csontos,~J.; Csóka,~J.; Szabó,~P.~B.; Gyevi-Nagy,~L.; Hégely,~B. \latin{et~al.}  The MRCC program system: Accurate quantum chemistry from water to proteins. \emph{The Journal of Chemical Physics} \textbf{2020}, \emph{152}, 074107\relax
\mciteBstWouldAddEndPuncttrue
\mciteSetBstMidEndSepPunct{\mcitedefaultmidpunct}
{\mcitedefaultendpunct}{\mcitedefaultseppunct}\relax
\EndOfBibitem
\bibitem[Kállay \latin{et~al.}()Kállay, Nagy, Mester, Gyevi-Nagy, Csóka, Szabó, Rolik, Samu, Hégely, Ladóczki, Petrov, Csontos, Ganyecz, Ladjánszki, Szegedy, Farkas, Mezei, Horváth, and Lőrincz]{mrcc}
Kállay,~M.; Nagy,~P.~R.; Mester,~D.; Gyevi-Nagy,~L.; Csóka,~J.; Szabó,~P.~B.; Rolik,~Z.; Samu,~G.; Hégely,~B.; Ladóczki,~B. \latin{et~al.}  MRCC: A quantum chemical program suite. \url{http://www.mrcc.hu}\relax
\mciteBstWouldAddEndPuncttrue
\mciteSetBstMidEndSepPunct{\mcitedefaultmidpunct}
{\mcitedefaultendpunct}{\mcitedefaultseppunct}\relax
\EndOfBibitem
\bibitem[Smith \latin{et~al.}()Smith, Roitberg, and Isayev]{smith2018transforming}
Smith,~J.~S.; Roitberg,~A.~E.; Isayev,~O. Transforming {{Computational Drug Discovery}} with {{Machine Learning}} and {{AI}}. \emph{9}, 1065--1069\relax
\mciteBstWouldAddEndPuncttrue
\mciteSetBstMidEndSepPunct{\mcitedefaultmidpunct}
{\mcitedefaultendpunct}{\mcitedefaultseppunct}\relax
\EndOfBibitem
\bibitem[Baiardi \latin{et~al.}()Baiardi, Grimmel, Steiner, Türtscher, Unsleber, Weymuth, and Reiher]{baiardi2022expansive}
Baiardi,~A.; Grimmel,~S.~A.; Steiner,~M.; Türtscher,~P.~L.; Unsleber,~J.~P.; Weymuth,~T.; Reiher,~M. Expansive {{Quantum Mechanical Exploration}} of {{Chemical Reaction Paths}}. \emph{55}, 35--43\relax
\mciteBstWouldAddEndPuncttrue
\mciteSetBstMidEndSepPunct{\mcitedefaultmidpunct}
{\mcitedefaultendpunct}{\mcitedefaultseppunct}\relax
\EndOfBibitem
\bibitem[Verma \latin{et~al.}()Verma, Kumar, Kumar, Ray, and Khandelwal]{verma2023application}
Verma,~G.; Kumar,~B.; Kumar,~C.; Ray,~A.; Khandelwal,~M. Application of {{KRR}}, {{K-NN}} and {{GPR Algorithms}} for {{Predicting}} the {{Soaked CBR}} of {{Fine-Grained Plastic Soils}}. \emph{48}, 13901--13927\relax
\mciteBstWouldAddEndPuncttrue
\mciteSetBstMidEndSepPunct{\mcitedefaultmidpunct}
{\mcitedefaultendpunct}{\mcitedefaultseppunct}\relax
\EndOfBibitem
\bibitem[Bortolussi \latin{et~al.}(2025)Bortolussi, Sandstr\"om, Partovi, Mikkil\"a, Rinke, and Rissanen]{Bortolussi25}
Bortolussi,~F.; Sandstr\"om,~H.; Partovi,~F.; Mikkil\"a,~J.; Rinke,~P.; Rissanen,~M. Technical note: Towards atmospheric compound identification in chemical ionization mass spectrometry with pesticide standards and machine learning. \emph{Atmospheric Chemistry and Physics} \textbf{2025}, \emph{25}, 685--704\relax
\mciteBstWouldAddEndPuncttrue
\mciteSetBstMidEndSepPunct{\mcitedefaultmidpunct}
{\mcitedefaultendpunct}{\mcitedefaultseppunct}\relax
\EndOfBibitem
\bibitem[Baird \latin{et~al.}(2023)Baird, Liu, Sayeed, and Sparks]{Baird23}
Baird,~S.~G.; Liu,~M.; Sayeed,~H.~M.; Sparks,~T.~D. In \emph{Comprehensive Inorganic Chemistry III (Third Edition)}, third edition ed.; Reedijk,~J., Poeppelmeier,~K.~R., Eds.; Elsevier: Oxford, 2023; pp 3--23\relax
\mciteBstWouldAddEndPuncttrue
\mciteSetBstMidEndSepPunct{\mcitedefaultmidpunct}
{\mcitedefaultendpunct}{\mcitedefaultseppunct}\relax
\EndOfBibitem
\bibitem[Pande \latin{et~al.}(2022)Pande, Shrivastava, Shilling, Zelenyuk, Zhang, Chen, Ng, Zhang, Takeuchi, Nah, Rasool, Zhang, Zhao, and Liu]{Paritosh22}
Pande,~P.; Shrivastava,~M.; Shilling,~J.~E.; Zelenyuk,~A.; Zhang,~Q.; Chen,~Q.; Ng,~N.~L.; Zhang,~Y.; Takeuchi,~M.; Nah,~T. \latin{et~al.}  Novel Application of Machine Learning Techniques for Rapid Source Apportionment of Aerosol Mass Spectrometer Datasets. \emph{ACS Earth and Space Chemistry} \textbf{2022}, \emph{6}, 932--942\relax
\mciteBstWouldAddEndPuncttrue
\mciteSetBstMidEndSepPunct{\mcitedefaultmidpunct}
{\mcitedefaultendpunct}{\mcitedefaultseppunct}\relax
\EndOfBibitem
\bibitem[Keith \latin{et~al.}(2021)Keith, Vassilev-Galindo, Cheng, Chmiela, Gastegger, Müller, and Tkatchenko]{keith21}
Keith,~J.~A.; Vassilev-Galindo,~V.; Cheng,~B.; Chmiela,~S.; Gastegger,~M.; Müller,~K.-R.; Tkatchenko,~A. Combining machine learning and computational chemistry for predictive insights into chemical systems. \emph{Chem. Rev.} \textbf{2021}, \emph{121}, 9816--9872\relax
\mciteBstWouldAddEndPuncttrue
\mciteSetBstMidEndSepPunct{\mcitedefaultmidpunct}
{\mcitedefaultendpunct}{\mcitedefaultseppunct}\relax
\EndOfBibitem
\bibitem[Meuwly(2021)]{meuwly21}
Meuwly,~M. Machine learning for chemical reactions. \emph{Chem. Rev.} \textbf{2021}, \emph{121}, 10218--10239\relax
\mciteBstWouldAddEndPuncttrue
\mciteSetBstMidEndSepPunct{\mcitedefaultmidpunct}
{\mcitedefaultendpunct}{\mcitedefaultseppunct}\relax
\EndOfBibitem
\bibitem[Kuntz and Wilson(2022)Kuntz, and Wilson]{kuntz22}
Kuntz,~D.; Wilson,~A.~K. Machine learning, artificial intelligence, and chemistry: How smart algorithms are reshaping simulation and the laboratory. \emph{Pure Appl. Chem} \textbf{2022}, \emph{94}, 1019--1054\relax
\mciteBstWouldAddEndPuncttrue
\mciteSetBstMidEndSepPunct{\mcitedefaultmidpunct}
{\mcitedefaultendpunct}{\mcitedefaultseppunct}\relax
\EndOfBibitem
\bibitem[Katritzky \latin{et~al.}(2010)Katritzky, Kuanar, Slavov, Hall, Karelson, Kahn, and Dobchev]{katritzky10}
Katritzky,~A.~R.; Kuanar,~M.; Slavov,~S.; Hall,~C.~D.; Karelson,~M.; Kahn,~I.; Dobchev,~D.~A. Quantitative correlation of physical and chemical properties with chemical structure: Utility for prediction. \emph{Chem. Rev.} \textbf{2010}, \emph{110}, 5714--5789\relax
\mciteBstWouldAddEndPuncttrue
\mciteSetBstMidEndSepPunct{\mcitedefaultmidpunct}
{\mcitedefaultendpunct}{\mcitedefaultseppunct}\relax
\EndOfBibitem
\bibitem[Anstine and Isayev()Anstine, and Isayev]{anstine2023machine}
Anstine,~D.~M.; Isayev,~O. Machine {{Learning Interatomic Potentials}} and {{Long-Range Physics}}. \emph{127}, 2417--2431\relax
\mciteBstWouldAddEndPuncttrue
\mciteSetBstMidEndSepPunct{\mcitedefaultmidpunct}
{\mcitedefaultendpunct}{\mcitedefaultseppunct}\relax
\EndOfBibitem
\bibitem[Sandstr\"om and Rinke(2025)Sandstr\"om, and Rinke]{sandstrom25}
Sandstr\"om,~H.; Rinke,~P. Similarity-based analysis of atmospheric organic compounds for machine learning applications. \emph{Geoscientific Model Development} \textbf{2025}, \emph{18}, 2701--2724\relax
\mciteBstWouldAddEndPuncttrue
\mciteSetBstMidEndSepPunct{\mcitedefaultmidpunct}
{\mcitedefaultendpunct}{\mcitedefaultseppunct}\relax
\EndOfBibitem
\bibitem[Lumiaro \latin{et~al.}(2021)Lumiaro, Todorovi\'c, Kurten, Vehkam\"aki, and Rinke]{lumiaro21}
Lumiaro,~E.; Todorovi\'c,~M.; Kurten,~T.; Vehkam\"aki,~H.; Rinke,~P. Predicting gas--particle partitioning coefficients of atmospheric molecules with machine learning. \emph{Atmos. Chem. Phys.} \textbf{2021}, \emph{21}, 13227--13246\relax
\mciteBstWouldAddEndPuncttrue
\mciteSetBstMidEndSepPunct{\mcitedefaultmidpunct}
{\mcitedefaultendpunct}{\mcitedefaultseppunct}\relax
\EndOfBibitem
\bibitem[Hyttinen \latin{et~al.}(2022)Hyttinen, Pihlajam{\" a}ki, and H{\" a}kkinen]{hyttinen22}
Hyttinen,~N.; Pihlajam{\" a}ki,~A.; H{\" a}kkinen,~H. Machine Learning for Predicting Chemical Potentials of Multifunctional Organic Compounds in Atmospherically Relevant Solutions. \emph{J. Phys. Chem. Letters} \textbf{2022}, \emph{13}, 9928--9933\relax
\mciteBstWouldAddEndPuncttrue
\mciteSetBstMidEndSepPunct{\mcitedefaultmidpunct}
{\mcitedefaultendpunct}{\mcitedefaultseppunct}\relax
\EndOfBibitem
\bibitem[Kr\"uger \latin{et~al.}(2025)Kr\"uger, Galeazzo, Eremets, Schmidt, P\"oschl, Shiraiwa, and Berkemeier]{kruger25}
Kr\"uger,~M.; Galeazzo,~T.; Eremets,~I.; Schmidt,~B.; P\"oschl,~U.; Shiraiwa,~M.; Berkemeier,~T. Improved vapor pressure predictions using group contribution-assisted graph convolutional neural networks (GC$^{2}$NN). \emph{EGUsphere} \textbf{2025}, \emph{2025}, 1--22\relax
\mciteBstWouldAddEndPuncttrue
\mciteSetBstMidEndSepPunct{\mcitedefaultmidpunct}
{\mcitedefaultendpunct}{\mcitedefaultseppunct}\relax
\EndOfBibitem
\bibitem[Kubečka \latin{et~al.}()Kubečka, Ayoubi, Tang, Knattrup, Engsvang, Wu, and Elm]{kubecka2024accurate}
Kubečka,~J.; Ayoubi,~D.; Tang,~Z.; Knattrup,~Y.; Engsvang,~M.; Wu,~H.; Elm,~J. Accurate Modeling of the Potential Energy Surface of Atmospheric Molecular Clusters Boosted by Neural Networks. \emph{3}, 1438--1451\relax
\mciteBstWouldAddEndPuncttrue
\mciteSetBstMidEndSepPunct{\mcitedefaultmidpunct}
{\mcitedefaultendpunct}{\mcitedefaultseppunct}\relax
\EndOfBibitem
\bibitem[Gupta \latin{et~al.}(2024)Gupta, Stulajter, Shaidu, Neaton, and de~Jong]{gupta24}
Gupta,~A.~K.; Stulajter,~M.~M.; Shaidu,~Y.; Neaton,~J.~B.; de~Jong,~W.~A. Equivariant Neural Networks Utilizing Molecular Clusters for Accurate Molecular Crystal Lattice Energy Predictions. \emph{ACS Omega} \textbf{2024}, \emph{9}, 40269--40282\relax
\mciteBstWouldAddEndPuncttrue
\mciteSetBstMidEndSepPunct{\mcitedefaultmidpunct}
{\mcitedefaultendpunct}{\mcitedefaultseppunct}\relax
\EndOfBibitem
\bibitem[Jiang \latin{et~al.}(2022)Jiang, Liu, Huang, Feng, Wang, Wang, Ge, Liu, Guang, and Huang]{jiang22}
Jiang,~S.; Liu,~Y.-R.; Huang,~T.; Feng,~Y.-J.; Wang,~C.-Y.; Wang,~Z.-Q.; Ge,~B.-J.; Liu,~Q.-S.; Guang,~W.-R.; Huang,~W. Towards fully ab initio simulation of atmospheric aerosol nucleation. \emph{Nat. Commun.} \textbf{2022}, \emph{13}, 6067\relax
\mciteBstWouldAddEndPuncttrue
\mciteSetBstMidEndSepPunct{\mcitedefaultmidpunct}
{\mcitedefaultendpunct}{\mcitedefaultseppunct}\relax
\EndOfBibitem
\bibitem[Jiang \latin{et~al.}(2023)Jiang, Liu, Wang, and Huang]{jiang23}
Jiang,~S.; Liu,~Y.-R.; Wang,~C.-Y.; Huang,~T. Benchmarking general neural network potential {ANI}-2x on aerosol nucleation molecular clusters. \emph{Int. J. Quantum Chem.} \textbf{2023}, \emph{123}, e27087\relax
\mciteBstWouldAddEndPuncttrue
\mciteSetBstMidEndSepPunct{\mcitedefaultmidpunct}
{\mcitedefaultendpunct}{\mcitedefaultseppunct}\relax
\EndOfBibitem
\bibitem[Ramakrishnan \latin{et~al.}(2015)Ramakrishnan, Dral, Rupp, and {von Lilienfeld}]{Ramakrishnan_jctc_2015}
Ramakrishnan,~R.; Dral,~P.~O.; Rupp,~M.; {von Lilienfeld},~O.~A. Big Data Meets Quantum Chemistry Approximations: The $\Delta$-Machine Learning Approach. \emph{J. Chem. Theory Comput.} \textbf{2015}, \emph{11}, 2087--2096\relax
\mciteBstWouldAddEndPuncttrue
\mciteSetBstMidEndSepPunct{\mcitedefaultmidpunct}
{\mcitedefaultendpunct}{\mcitedefaultseppunct}\relax
\EndOfBibitem
\bibitem[Kubečka \latin{et~al.}()Kubečka, Christensen, Rasmussen, and Elm]{kubecka2022Quantum}
Kubečka,~J.; Christensen,~A.~S.; Rasmussen,~F.~R.; Elm,~J. Quantum {{Machine Learning Approach}} for {{Studying Atmospheric Cluster Formation}}. \emph{9}, 239--244\relax
\mciteBstWouldAddEndPuncttrue
\mciteSetBstMidEndSepPunct{\mcitedefaultmidpunct}
{\mcitedefaultendpunct}{\mcitedefaultseppunct}\relax
\EndOfBibitem
\bibitem[Knattrup \latin{et~al.}()Knattrup, Kubečka, Ayoubi, and Elm]{knattrup2023clusterome}
Knattrup,~Y.; Kubečka,~J.; Ayoubi,~D.; Elm,~J. Clusterome: {{A Comprehensive Data Set}} of {{Atmospheric Molecular Clusters}} for {{Machine Learning Applications}}. \emph{8}, 25155--25164\relax
\mciteBstWouldAddEndPuncttrue
\mciteSetBstMidEndSepPunct{\mcitedefaultmidpunct}
{\mcitedefaultendpunct}{\mcitedefaultseppunct}\relax
\EndOfBibitem
\bibitem[Hastie \latin{et~al.}()Hastie, Tibshirani, and Friedman]{hastie2009}
Hastie,~T.; Tibshirani,~R.; Friedman,~J.~H. \emph{The Elements of Statistical Learning: Data Mining, Inference, and Prediction}, 2nd ed.; Springer Series in Statistics; Springer\relax
\mciteBstWouldAddEndPuncttrue
\mciteSetBstMidEndSepPunct{\mcitedefaultmidpunct}
{\mcitedefaultendpunct}{\mcitedefaultseppunct}\relax
\EndOfBibitem
\bibitem[Omohundro(1989)]{omohundro1989Five}
Omohundro,~S. \emph{Five Balltree Construction Algorithms}; 1989\relax
\mciteBstWouldAddEndPuncttrue
\mciteSetBstMidEndSepPunct{\mcitedefaultmidpunct}
{\mcitedefaultendpunct}{\mcitedefaultseppunct}\relax
\EndOfBibitem
\bibitem[Beyer \latin{et~al.}()Beyer, Goldstein, Ramakrishnan, and Shaft]{beyer1999meaningful}
Beyer,~K.; Goldstein,~J.; Ramakrishnan,~R.; Shaft,~U. When Is “Nearest Neighbor” Meaningful? Database Theory — {{ICDT}}'99. pp 217--235\relax
\mciteBstWouldAddEndPuncttrue
\mciteSetBstMidEndSepPunct{\mcitedefaultmidpunct}
{\mcitedefaultendpunct}{\mcitedefaultseppunct}\relax
\EndOfBibitem
\bibitem[Murphy()]{murphy2022probabilistic}
Murphy,~K.~P. \emph{Probabilistic Machine Learning: {{An}} Introduction}; MIT Press\relax
\mciteBstWouldAddEndPuncttrue
\mciteSetBstMidEndSepPunct{\mcitedefaultmidpunct}
{\mcitedefaultendpunct}{\mcitedefaultseppunct}\relax
\EndOfBibitem
\bibitem[Janet and Kulik()Janet, and Kulik]{janet2020Machinea}
Janet,~J.~P.; Kulik,~H.~J. \emph{Machine {{Learning}} in {{Chemistry}}}; {{ACS In Focus}}; American Chemical Society\relax
\mciteBstWouldAddEndPuncttrue
\mciteSetBstMidEndSepPunct{\mcitedefaultmidpunct}
{\mcitedefaultendpunct}{\mcitedefaultseppunct}\relax
\EndOfBibitem
\bibitem[Golub and Van~Loan()Golub, and Van~Loan]{golub2013matrix}
Golub,~G.~H.; Van~Loan,~C.~F. \emph{Matrix {{Computations}}}; Johns Hopkins University Press\relax
\mciteBstWouldAddEndPuncttrue
\mciteSetBstMidEndSepPunct{\mcitedefaultmidpunct}
{\mcitedefaultendpunct}{\mcitedefaultseppunct}\relax
\EndOfBibitem
\bibitem[Coppersmith and Winograd()Coppersmith, and Winograd]{coppersmith1990Matrix}
Coppersmith,~D.; Winograd,~S. Matrix Multiplication via Arithmetic Progressions. \emph{9}, 251--280\relax
\mciteBstWouldAddEndPuncttrue
\mciteSetBstMidEndSepPunct{\mcitedefaultmidpunct}
{\mcitedefaultendpunct}{\mcitedefaultseppunct}\relax
\EndOfBibitem
\bibitem[Langer \latin{et~al.}()Langer, Goeßmann, and Rupp]{langer2022Representations}
Langer,~M.~F.; Goeßmann,~A.; Rupp,~M. Representations of Molecules and Materials for Interpolation of Quantum-Mechanical Simulations via Machine Learning. \emph{8}, 41\relax
\mciteBstWouldAddEndPuncttrue
\mciteSetBstMidEndSepPunct{\mcitedefaultmidpunct}
{\mcitedefaultendpunct}{\mcitedefaultseppunct}\relax
\EndOfBibitem
\bibitem[Ramakrishnan \latin{et~al.}()Ramakrishnan, Dral, Rupp, and Lilienfeld]{ramakrishnan2015big}
Ramakrishnan,~R.; Dral,~P.~O.; Rupp,~M.; Lilienfeld,~O. Big {{Data Meets Quantum Chemistry Approximations}}: {The $\Delta$-Machine Learning Approach}. \emph{11}, 2087--2096\relax
\mciteBstWouldAddEndPuncttrue
\mciteSetBstMidEndSepPunct{\mcitedefaultmidpunct}
{\mcitedefaultendpunct}{\mcitedefaultseppunct}\relax
\EndOfBibitem
\bibitem[Phillips and Venkatasubramanian()Phillips, and Venkatasubramanian]{phillips2011gentle}
Phillips,~J.~M.; Venkatasubramanian,~S. A Gentle Introduction to the Kernel Distance. \url{https://arxiv.org/abs/1103.1625}\relax
\mciteBstWouldAddEndPuncttrue
\mciteSetBstMidEndSepPunct{\mcitedefaultmidpunct}
{\mcitedefaultendpunct}{\mcitedefaultseppunct}\relax
\EndOfBibitem
\bibitem[Faber \latin{et~al.}()Faber, Christensen, Huang, and Anatole~von Lilienfeld]{faber2018alchemical}
Faber,~F.~A.; Christensen,~A.~S.; Huang,~B.; Anatole~von Lilienfeld,~O. Alchemical and Structural Distribution Based Representation for Universal Quantum Machine Learning. \emph{148}, 241717\relax
\mciteBstWouldAddEndPuncttrue
\mciteSetBstMidEndSepPunct{\mcitedefaultmidpunct}
{\mcitedefaultendpunct}{\mcitedefaultseppunct}\relax
\EndOfBibitem
\bibitem[Weinberger and Tesauro()Weinberger, and Tesauro]{weinberger2007metric}
Weinberger,~K.~Q.; Tesauro,~G. Metric Learning for Kernel Regression. Proceedings of the Eleventh International Conference on Artificial Intelligence and Statistics. pp 612--619\relax
\mciteBstWouldAddEndPuncttrue
\mciteSetBstMidEndSepPunct{\mcitedefaultmidpunct}
{\mcitedefaultendpunct}{\mcitedefaultseppunct}\relax
\EndOfBibitem
\bibitem[Kanagawa()]{kanagawa2024fast}
Kanagawa,~M. Fast Computation of Leave-One-out Cross-Validation for {{k}}-{{NN}} Regression. \url{https://arxiv.org/abs/2405.04919}\relax
\mciteBstWouldAddEndPuncttrue
\mciteSetBstMidEndSepPunct{\mcitedefaultmidpunct}
{\mcitedefaultendpunct}{\mcitedefaultseppunct}\relax
\EndOfBibitem
\bibitem[Efron and Tibshirani()Efron, and Tibshirani]{efron1993introduction}
Efron,~B.; Tibshirani,~R. \emph{An Introduction to the Bootstrap}; Monographs on Statistics and Applied Probability 57; Chapman \& Hall\relax
\mciteBstWouldAddEndPuncttrue
\mciteSetBstMidEndSepPunct{\mcitedefaultmidpunct}
{\mcitedefaultendpunct}{\mcitedefaultseppunct}\relax
\EndOfBibitem
\bibitem[Efron and Stein(1981)Efron, and Stein]{efron1982jackknife}
Efron,~B.; Stein,~C. The Jackknife Estimate of Variance. \emph{The Annals of Statistics} \textbf{1981}, \emph{9}, 586--596\relax
\mciteBstWouldAddEndPuncttrue
\mciteSetBstMidEndSepPunct{\mcitedefaultmidpunct}
{\mcitedefaultendpunct}{\mcitedefaultseppunct}\relax
\EndOfBibitem
\bibitem[Musil \latin{et~al.}(2021)Musil, Grisafi, Bartók, Ortner, Csányi, and Ceriotti]{musil2021physics}
Musil,~F.; Grisafi,~A.; Bartók,~A.~P.; Ortner,~C.; Csányi,~G.; Ceriotti,~M. Physics-Inspired Structural Representations for Molecules and Materials. \emph{Chemical Reviews} \textbf{2021}, \emph{121}, 9759--9815\relax
\mciteBstWouldAddEndPuncttrue
\mciteSetBstMidEndSepPunct{\mcitedefaultmidpunct}
{\mcitedefaultendpunct}{\mcitedefaultseppunct}\relax
\EndOfBibitem
\bibitem[Christensen \latin{et~al.}()Christensen, Bratholm, Faber, and Anatole~von Lilienfeld]{christensen2020FCHL}
Christensen,~A.~S.; Bratholm,~L.~A.; Faber,~F.~A.; Anatole~von Lilienfeld,~O. {{FCHL}} Revisited: {{Faster}} and More Accurate Quantum Machine Learning. \emph{152}, 044107\relax
\mciteBstWouldAddEndPuncttrue
\mciteSetBstMidEndSepPunct{\mcitedefaultmidpunct}
{\mcitedefaultendpunct}{\mcitedefaultseppunct}\relax
\EndOfBibitem
\bibitem[Kubečka \latin{et~al.}()Kubečka, Besel, Kurtén, Myllys, and Vehkamäki]{kubecka2019configurational}
Kubečka,~J.; Besel,~V.; Kurtén,~T.; Myllys,~N.; Vehkamäki,~H. Configurational Sampling of Noncovalent (Atmospheric) Molecular Clusters: {{Sulfuric}} Acid and Guanidine. \emph{123}, 6022--6033\relax
\mciteBstWouldAddEndPuncttrue
\mciteSetBstMidEndSepPunct{\mcitedefaultmidpunct}
{\mcitedefaultendpunct}{\mcitedefaultseppunct}\relax
\EndOfBibitem
\bibitem[Kubečka \latin{et~al.}(2023)Kubečka, Besel, Neefjes, Knattrup, Kurtén, Vehkamäki, and Elm]{kubecka23jkcs}
Kubečka,~J.; Besel,~V.; Neefjes,~I.; Knattrup,~Y.; Kurtén,~T.; Vehkamäki,~H.; Elm,~J. Computational tools for handling molecular clusters: Configurational sampling, storage, analysis, and machine learning. \emph{ACS Omega} \textbf{2023}, \emph{8}, 45115--45128\relax
\mciteBstWouldAddEndPuncttrue
\mciteSetBstMidEndSepPunct{\mcitedefaultmidpunct}
{\mcitedefaultendpunct}{\mcitedefaultseppunct}\relax
\EndOfBibitem
\bibitem[Ramakrishnan \latin{et~al.}()Ramakrishnan, Dral, Rupp, and Lilienfeld]{ramakrishnan2014qm9}
Ramakrishnan,~R.; Dral,~P.~O.; Rupp,~M.; Lilienfeld,~O. Quantum Chemistry Structures and Properties of 134 Kilo Molecules. \emph{1}, 140022\relax
\mciteBstWouldAddEndPuncttrue
\mciteSetBstMidEndSepPunct{\mcitedefaultmidpunct}
{\mcitedefaultendpunct}{\mcitedefaultseppunct}\relax
\EndOfBibitem
\bibitem[Paasonen \latin{et~al.}()Paasonen, Nieminen, Asmi, Manninen, Petäjä, Plass-Dülmer, Flentje, Birmili, Wiedensohler, Hõrrak, Metzger, Hamed, Laaksonen, Facchini, Kerminen, and Kulmala]{paasonen2010roles}
Paasonen,~P.; Nieminen,~T.; Asmi,~E.; Manninen,~H.~E.; Petäjä,~T.; Plass-Dülmer,~C.; Flentje,~H.; Birmili,~W.; Wiedensohler,~A.; Hõrrak,~U. \latin{et~al.}  On the Roles of Sulphuric Acid and Low-Volatility Organic Vapours in the Initial Steps of Atmospheric New Particle Formation. \emph{10}, 11223--11242\relax
\mciteBstWouldAddEndPuncttrue
\mciteSetBstMidEndSepPunct{\mcitedefaultmidpunct}
{\mcitedefaultendpunct}{\mcitedefaultseppunct}\relax
\EndOfBibitem
\bibitem[Kube{\v c}ka \latin{et~al.}(2022)Kube{\v c}ka, Christensen, Rasmussen, and Elm]{Kubecka_estlett_2022}
Kube{\v c}ka,~J.; Christensen,~A.~S.; Rasmussen,~F.~R.; Elm,~J. Quantum Machine Learning Approach for Studying Atmospheric Cluster Formation. \emph{Environ. Sci. Technol. Lett.} \textbf{2022}, \emph{9}, 239--244\relax
\mciteBstWouldAddEndPuncttrue
\mciteSetBstMidEndSepPunct{\mcitedefaultmidpunct}
{\mcitedefaultendpunct}{\mcitedefaultseppunct}\relax
\EndOfBibitem
\bibitem[Elm \latin{et~al.}(2013)Elm, Bilde, and Mikkelsen]{Elm2013}
Elm,~J.; Bilde,~M.; Mikkelsen,~K.~V. Assessment of binding energies of atmospherically relevant clusters. \emph{Phys. Chem. Chem. Phys.} \textbf{2013}, \emph{15}, 16442--16445\relax
\mciteBstWouldAddEndPuncttrue
\mciteSetBstMidEndSepPunct{\mcitedefaultmidpunct}
{\mcitedefaultendpunct}{\mcitedefaultseppunct}\relax
\EndOfBibitem
\bibitem[Elm and Kristensen(2017)Elm, and Kristensen]{Elm2017}
Elm,~J.; Kristensen,~K. Basis set convergence of the binding energies of strongly hydrogen-bonded atmospheric clusters. \emph{Phys. Chem. Chem. Phys.} \textbf{2017}, \emph{19}, 1122--1133\relax
\mciteBstWouldAddEndPuncttrue
\mciteSetBstMidEndSepPunct{\mcitedefaultmidpunct}
{\mcitedefaultendpunct}{\mcitedefaultseppunct}\relax
\EndOfBibitem
\bibitem[Schmitz and Elm(2020)Schmitz, and Elm]{schmitz2020}
Schmitz,~G.; Elm,~J. Assessment of the DLPNO binding energies of strongly noncovalent bonded atmospheric molecular clusters. \emph{ACS Omega} \textbf{2020}, \emph{5}, 7601--7612\relax
\mciteBstWouldAddEndPuncttrue
\mciteSetBstMidEndSepPunct{\mcitedefaultmidpunct}
{\mcitedefaultendpunct}{\mcitedefaultseppunct}\relax
\EndOfBibitem
\bibitem[Jensen \latin{et~al.}(2022)Jensen, Kubečka, Schmitz, Christiansen, and Elm]{Jensen2022}
Jensen,~A.~B.; Kubečka,~J.; Schmitz,~G.; Christiansen,~O.; Elm,~J. Massive assessment of the binding energies of atmospheric molecular clusters. \emph{J. Chem. Theory Comput.} \textbf{2022}, \emph{18}, 7373--7383\relax
\mciteBstWouldAddEndPuncttrue
\mciteSetBstMidEndSepPunct{\mcitedefaultmidpunct}
{\mcitedefaultendpunct}{\mcitedefaultseppunct}\relax
\EndOfBibitem
\bibitem[Elm(2021)]{Clusteromics_I}
Elm,~J. Clusteromics I: Principles, Protocols and Applications to Sulfuric Acid - Base Cluster Formation. \emph{ACS Omega} \textbf{2021}, \emph{6}, 7804--7814\relax
\mciteBstWouldAddEndPuncttrue
\mciteSetBstMidEndSepPunct{\mcitedefaultmidpunct}
{\mcitedefaultendpunct}{\mcitedefaultseppunct}\relax
\EndOfBibitem
\bibitem[Elm(2021)]{Clusteromics_II}
Elm,~J. Clusteromics II: Methanesulfonic Acid-Base Cluster Formation. \emph{ACS Omega} \textbf{2021}, \emph{6}, 17035--17044\relax
\mciteBstWouldAddEndPuncttrue
\mciteSetBstMidEndSepPunct{\mcitedefaultmidpunct}
{\mcitedefaultendpunct}{\mcitedefaultseppunct}\relax
\EndOfBibitem
\bibitem[Elm(2022)]{Clusteromics_III}
Elm,~J. Clusteromics III: Acid Synergy in Sulfuric Acid-Methanesulfonic Acid-Base Cluster Formation. \emph{ACS Omega} \textbf{2022}, \emph{7}, 15206--15214\relax
\mciteBstWouldAddEndPuncttrue
\mciteSetBstMidEndSepPunct{\mcitedefaultmidpunct}
{\mcitedefaultendpunct}{\mcitedefaultseppunct}\relax
\EndOfBibitem
\bibitem[Knattrup and Elm(2022)Knattrup, and Elm]{Clusteromics_IV}
Knattrup,~Y.; Elm,~J. Clusteromics {IV}: The Role of Nitric Acid in Atmospheric Cluster Formation. \emph{{ACS} Omega} \textbf{2022}, \emph{7}, 31551--31560\relax
\mciteBstWouldAddEndPuncttrue
\mciteSetBstMidEndSepPunct{\mcitedefaultmidpunct}
{\mcitedefaultendpunct}{\mcitedefaultseppunct}\relax
\EndOfBibitem
\bibitem[Ayoubi \latin{et~al.}(2023)Ayoubi, Knattrup, and Elm]{Clusteromics_V}
Ayoubi,~D.; Knattrup,~Y.; Elm,~J. Clusteromics V: Organic Enhanced Atmospheric Cluster Formation. \emph{ACS Omega} \textbf{2023}, \emph{8}, 9621--9629\relax
\mciteBstWouldAddEndPuncttrue
\mciteSetBstMidEndSepPunct{\mcitedefaultmidpunct}
{\mcitedefaultendpunct}{\mcitedefaultseppunct}\relax
\EndOfBibitem
\bibitem[Rupp \latin{et~al.}(2012)Rupp, Tkatchenko, M\"uller, and von Lilienfeld]{rupp2012fast}
Rupp,~M.; Tkatchenko,~A.; M\"uller,~K.-R.; von Lilienfeld,~O.~A. Fast and Accurate Modeling of Molecular Atomization Energies with Machine Learning. \emph{Phys. Rev. Lett.} \textbf{2012}, \emph{108}, 058301\relax
\mciteBstWouldAddEndPuncttrue
\mciteSetBstMidEndSepPunct{\mcitedefaultmidpunct}
{\mcitedefaultendpunct}{\mcitedefaultseppunct}\relax
\EndOfBibitem
\bibitem[Hansen \latin{et~al.}()Hansen, Biegler, Ramakrishnan, Pronobis, von Lilienfeld, Müller, and Tkatchenko]{hansen2015machine}
Hansen,~K.; Biegler,~F.; Ramakrishnan,~R.; Pronobis,~W.; von Lilienfeld,~O.~A.; Müller,~K.-R.; Tkatchenko,~A. Machine {{Learning Predictions}} of {{Molecular Properties}}: {{Accurate Many-Body Potentials}} and {{Nonlocality}} in {{Chemical Space}}. \emph{6}, 2326--2331\relax
\mciteBstWouldAddEndPuncttrue
\mciteSetBstMidEndSepPunct{\mcitedefaultmidpunct}
{\mcitedefaultendpunct}{\mcitedefaultseppunct}\relax
\EndOfBibitem
\bibitem[Khan \latin{et~al.}()Khan, Heinen, and Anatole~von Lilienfeld]{khan2023mbdf}
Khan,~D.; Heinen,~S.; Anatole~von Lilienfeld,~O. Kernel Based Quantum Machine Learning at Record Rate: {{Many-body}} Distribution Functionals as Compact Representations. \emph{159}, 034106\relax
\mciteBstWouldAddEndPuncttrue
\mciteSetBstMidEndSepPunct{\mcitedefaultmidpunct}
{\mcitedefaultendpunct}{\mcitedefaultseppunct}\relax
\EndOfBibitem
\end{mcitethebibliography}

\newpage
\appendix

\section{Molecular representation comparison}
\label{supsec:representations}

The choice of representation is a crucial step in applying ML on chemical data.
While the FCHL representation was introduced in the main body of the text, we also included the Coulomb matrix (CM), Bag-of-Bonds (BoB) and many-body functional distributions (MBDF) representations in our analysis for completeness.

Coulomb matrices \cite{rupp2012fast} are a simple yet effective molecular representation.
The representation is global and aims to mimic the electrostatic interaction between nuclei in a chemical system.
Bag-of-bonds \cite{hansen2015machine} takes inspiration from the bag-of-words representations common in natural language processing in computer science where the electrostatic interactions between nuclei are bagged to produce a single vector as the representation.
Similar to FCHL19, the MBDF \cite{khan2023mbdf} representation is designed to minimise the size of the representation without dramatically impacting predictive accuracy.
Furthermore, both representations aim to capture the distribution of the many-body properties of the chemical system.
While FCHL encodes each of these features explicitly, MBDF instead expresses the system through a finite set of integrals that describe these distributions.
Conceptually, FCHL provides a high-dimensional, explicit encoding of the local atomic environments by modelling the distributions of elements, distances, and angles.
In contrast, MBDF summarizes these distributions by computing statistical moments (e.g., mean, variance) to produce a lower-dimensional representation.

Table \ref{tbl:representations} shows the test set mean absolute error and train and test CPU times of $k$-NN models trained on $13,000$ samples of $\Delta$-learning data from SA--W clusters (identical to Section 4.2.1 in the main paper) and with various representations and metric types.
As we have alluded in previous sections, FCHL19 combined with the MLKR metric learning offers good accuracy with reasonable training time.
MBDF, on the other hand, does not perform as well and produces results similar to the much simpler bag-of-bonds.

While plain FCHL18 is unsuitable to be used with Euclidean or MLKR-based $k$-NN models due the representation being a 4D tensor, we did experiment with kernel-induced distance using FCHL18 as well.
As the table shows, the FCHL18-based kernel-induced distance $k$-NN achieves similar results to FCHL19 (i.e., nearly within one standard deviation), but is much slower due to the more complicated calculation of the kernel elements in both training and testing.

\begin{table}[h]
\centering
\caption{Comparison of representations and metric types in $k$-NN model applied to the SA--W dataset. Smaller values are better in each column, with best values highlighted in bold. The representations include bag-of-bonds (BoB), Coulomb matrices (CM), both FCHL18 and FCHL19 variants and MBDF. The metric types include Euclidean distance, MLKR-based learned metric and for FCHL18 and FCHL19, kernel-induced distance (denoted ``Kernel'' in the table).}
\label{tbl:representations}
\resizebox{\linewidth}{!}{
\begin{tabular}{lllll}
\toprule
     &      &         Test MAE (kcal/mol) &          Train CPU time (s) &           Test CPU time (s) \\
Representation & Metric type &                             &                             &                             \\
\midrule
BoB & Euclidean &             3.28 $\pm$ 0.05 &             0.01 $\pm$ 0.00 &             3.04 $\pm$ 0.04 \\
     & MLKR &             2.25 $\pm$ 0.04 &       10029.99 $\pm$ 759.65 &        2302.37 $\pm$ 789.87 \\
CM & Euclidean &             3.78 $\pm$ 0.07 &             0.01 $\pm$ 0.00 &             2.61 $\pm$ 0.01 \\
     & MLKR &             3.81 $\pm$ 0.07 &      13253.95 $\pm$ 1279.53 &        2001.73 $\pm$ 292.70 \\
FCHL18 & Kernel &             2.89 $\pm$ 0.05 &        9591.09 $\pm$ 215.50 &       65901.09 $\pm$ 625.11 \\
FCHL19 & Euclidean &             3.31 $\pm$ 0.06 &             0.01 $\pm$ 0.00 &  \bfseries{1.46 $\pm$ 0.01} \\
     & Kernel &             2.95 $\pm$ 0.05 &          361.99 $\pm$ 47.88 &      13914.71 $\pm$ 1202.50 \\
     & MLKR &  \bfseries{1.19 $\pm$ 0.02} &      14392.87 $\pm$ 1855.89 &        1908.42 $\pm$ 331.91 \\
MBDF & Euclidean &             2.92 $\pm$ 0.08 &  \bfseries{0.00 $\pm$ 0.00} &             1.51 $\pm$ 0.09 \\
     & MLKR &             2.23 $\pm$ 0.03 &       13640.22 $\pm$ 277.35 &        1658.75 $\pm$ 225.30 \\
\bottomrule
\end{tabular}
}
\end{table}
\end{document}